\documentclass[twoside,11pt]{article}

\usepackage{jmlr2e}

\usepackage{graphicx}
\usepackage{epstopdf}
\usepackage{enumerate}
\usepackage{multicol}
\usepackage{algorithmic}
\usepackage[lined,boxed,commentsnumbered,linesnumbered,ruled]{algorithm2e}

\usepackage{multirow}
\usepackage{color}
\usepackage{gensymb}

\SetKwInput{KwInput}{Input}
\SetKwInput{KwOutput}{Output}

\makeatletter
\newcommand*{\indep}{%
  \mathbin{%
    \mathpalette{\@indep}{}%
  }%
}
\newcommand*{\nindep}{%
  \mathbin{
    \mathpalette{\@indep}{\not}
  }%
}
\newcommand*{\@indep}[2]{%
  \sbox0{$#1\perp\m@th$}
  \sbox2{$#1=$}
  \sbox4{$#1\vcenter{}$}
  \rlap{\copy0}
  \dimen@=\dimexpr\ht2-\ht4-.2pt\relax
  \kern\dimen@
  {#2}%
  \kern\dimen@
  \copy0 
} 

\makeatother

\newcommand*{\QEDA}{\hfill\ensuremath{\square}}
\def\boxend{\hspace*{\fill} $\QEDA$}

\begin{document}
\begin{sloppy}

\title{Discovering Markov Blanket from Multiple interventional Datasets}

\author{\name Kui Yu \email Kui.Yu@unisa.edu.au \\
         \name Lin Liu \email Lin.Liu@unisa.edu.au \\
          \name Jiuyong Li \email Jiuyong.Li@unisa.edu.au \\
       \addr School of Information Technology and Mathematical Sciences\\
      University of South Australia\\
       Adelaide, 5095, SA, Australia
       }
       
\editor{}

\maketitle

\begin{abstract}
In this paper, we study the problem of discovering the Markov blanket (MB) of a target variable from multiple interventional datasets. Datasets attained from interventional experiments contain richer causal information than passively observed data (observational data) for MB discovery. However, almost all existing MB discovery methods are designed for finding MBs from a single observational dataset.  To identify MBs from multiple interventional datasets, we face two challenges: (1) unknown intervention variables; (2) nonidentical data distributions. To tackle the challenges, we theoretically analyze (a) under what conditions we can find the correct MB of a target variable, and (b) under what conditions we can identify the causes of the target variable via discovering its MB. Based on the theoretical analysis, we propose a new algorithm for discovering MBs from multiple interventional datasets, and present the conditions/assumptions which assure the correctness of the algorithm. To our knowledge, this work is the first to present the theoretical analyses about the conditions for MB discovery in multiple interventional datasets and the algorithm to find the MBs in relation to the conditions. Using benchmark Bayesian networks and real-world datasets, the experiments have validated the effectiveness and efficiency of the proposed algorithm in the paper.
\end{abstract}


\section{Introduction}\label{sec1}

The Markov blanket (MB) of a variable comprises its parents (direct causes), children (direct effects), and spouses (direct causes of children) in a causal Bayesian network, where the causal relationships among the set of variables under consideration are represented using a causal DAG (Directed Acyclic Graph)~\citep{pearl2009causality,spirtes2000causation}. That is, nodes of the causal DAG represent the variables and an edge $X\rightarrow Y$ indicates that $X$ is a direct cause of $Y$. As shown in Figure~\ref{fig1}(a), the MB of variable $T$ contains A (parent), B (child) and F (spouse). The MB of a variable provides a complete picture of the local causal structure around the variable, and thus learning MBs plays an essential role in local causal discovery~\citep{aliferis2010local1}. Moreover, it is well recognized that learning a causal DAG is computationally infeasible for a large number of variables~\citep{margaritis1999bayesian}, but if we can get the MBs of the variables, we are able to use them as constraints to reduce search spaces in the design of scalable local-to-global structure learning methods~\citep{tsamardinos2006max,aliferis2010local2}.

However almost all existing methods are for finding MBs from a single observational dataset and cannot distinguish causes from effects in a found MB, as it is only possible to identify the Markov equivalence class of a causal structure based on observational data~\citep{spirtes2000causation}. For example, the three structures, $A\rightarrow T\rightarrow B$, $A\leftarrow T\leftarrow B$, and $A\leftarrow T\rightarrow B$ all encode the same conditional independence statement, $A$ and $B$ are independent given $T$. All existing MB mining algorithms can only find the in(dependent) relations among $A$, $B$, and $T$, that is, the skeleton $A-T-B$ with the directions of the edges left unidentified.
\begin{figure}
\small
\centering
\includegraphics[height=1.6in,width=3.1in]{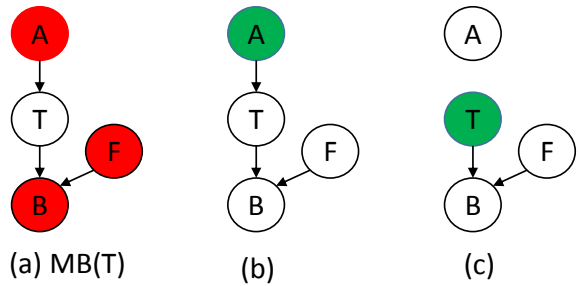}
\caption{(a) An example DAG showing the underlying causal relations among variables $T$, $A$, $B$ and $F$, (b) the post-intervention DAG as $A$ has been intervened ($A$ and $T$ are still dependent),  and (c) the post-intervention DAG as $T$ has been intervened ($A$ and $T$ become independent). Given the MB set of $T$, based on (b) and (c), we can infer $A$ is the parent of $T$~\citep{he2016causal}.}
\label{fig1}
\end{figure}

To distinguish between the structures above, an effective way is to use multiple interventional (experimental) datasets. 
For example, assume that the DAG in Figure~\ref{fig1} (a) shows the underlying (true) causal relations between variables $A$, $B$, $F$ and $T$.  As $A$ is a direct cause of $T$, $T$'s values will be affected by the change/manipulation of $A$, but the values of $A$ are not affected by the change of $T$. Now suppose that we have a dataset where $A$ is manipulated and another dataset where $T$ is manipulated, then from the first dataset, we should be able to learn that $A$ and $T$ are dependent and from the second dataset, $A$ and $T$ are independent (as indicated by the post-intervention DAGs shown in Figure~\ref{fig1} (b) and Figure~\ref{fig1} (c) respectively). Thus we can conclude from the results that $A$ is a cause of $T$, instead of the other way around~\citep{pearl2009causality}. 

From the example we see that interventional data contains rich causal information, and multiple interventional datasets, when used together, may help with the discovery of MBs, including the orientations of relations in MBs.

It is desirable to utilize the data in MB discovery, since there has been an increasing availability of interventional data collected from various sources, such as gene knockdown experiments by different labs for studying the same diseases~\citep{bareinboim2016}.

However, the challenge is that  in practice, we often do not know exactly which variables in the interventional datasets were manipulated. Question then arises as to when it is possible to make use of multiple interventional datasets to discover MBs, and furthermore to distinguish causes from effects in a found MB (which existing MB discovery methods fail to achieve). This leads to the following more specific questions:
\begin{itemize}
\item Under what manipulation settings (i.e. which variables were manipulated) can we discover the true MB of a target variable? 
\item Under what manipulation settings can we identify the causes of the target variable via finding its MB? 
\item How can we effectively find MBs from multiple interventional datasets? Multiple interventional datasets are not identically distributed since they are generated under different interventions. Thus we cannot simply pool all datasets together to find MBs. 
\end{itemize}


Although several algorithms have been proposed for mining causal structures from multiple interventional datasets~\citep{eberhardt2006n,hauser2012characterization,triantafillou2015constraint,he2016causal}, these methods are all for learning a global causal structure. As far as we know, there has been no method developed for discovering MBs from multiple interventional datasets, and almost all existing methods are for finding MBs from a single observational dataset. One may use those global causal structure learning algorithms for finding MBs from multiple interventional datasets, but as they are designed to mine the entire causal structures involving all variables in data, those algorithms are computationally intractable or suboptimal for MB discovery.
Moreover, in practice, it is often unnecessary and wasteful to find the entire structures as we are only interested in the local causal structure around one target variable.

Therefore methods specifically designed for discovering MBs from multiple interventional datasets are of demand. This paper is aimed at answering the questions mentioned above and our main contributions are as follows:
\begin{itemize}
\item  Given a target variable, we theoretically analyze: (1) under what variable manipulation settings we can find the correct MB of the target variable or not, and (2) under what variable manipulation settings, in a found MB, we can identify the causes (parents) of the target variable, and thus to distinguish among the target's parents, children and spouses.
\item Based on the theoretical analyses, we propose a new algorithm for the MB discovery with multiple interventional datasets, present the conditions/assumptions which assure the correctness of the algorithm, and validate the effectiveness and efficiency of the algorithm using benchmark Bayesian networks and real-world datasets.
\end{itemize}

The paper is organized as follows. Section~\ref{sec2} reviews the related work, and Section~\ref{sec3} gives notations and problem definition. Section~\ref{sec4} presents the theoretical analysis, while Section~\ref{sec5} proposes our new algorithm. Section~\ref{sec6} describes and discusses the experiments and Section~\ref{sec7} concludes the paper and presents future work.

\section{Related work}\label{sec2}

Many algorithms have been proposed for discovering MBs, but almost all the methods use a single observational dataset. Margaritis and Thrun~\citep{margaritis1999bayesian} proposed the GSMB algorithm, the first provably correct algorithm under the faithfulness assumption. Later, several variants for improving the reliability of GSMB, like IAMB~\citep{tsamardinos2003algorithms}, Inter-IAMB~\citep{tsamardinos2003algorithms}, and Fast-IAMB~\citep{yaramakala2005speculative}, HITON-MB~\citep{aliferis2010local1}, MMMB~\citep{tsamardinos2006max}, PCMB~\citep{pena2007towards}, and IPCMB~\citep{fu2008fast} were presented. 

Using the discovered MBs, many local-to-global structure learning~\citep{aliferis2010local2,pellet2008using,tsamardinos2006max,gao2017local,yang2016partial} and  local causal discovery~\citep{gao2015local,yin2008partial} approaches have been proposed for learning a global causal structure involving hundreds of variables and for discovering a local causal structure of a target variable.

For learning a global causal structure from multiple datasets, the first group of algorithms focuses on learning a joint maximal ancestral graph (MAG) from multiple observational datasets with overlapping variables, such as the SLPR algorithm~\citep{danks2002learning}, the ION algorithm~\citep{danks2009integrating}, the IOD algorithm~\citep{tillman2011learning}, and the INCA framework~\citep{tsamardinos2012towards}.

The second group of algorithms mines causal structures using both observational data and experimental data. These algorithms firstly mine a Markov equivalence class of an underlying DAG using observational datasets, and thus this may leave many edge directions undermined. Then, the methods conduct variable intervention experiments to orient the edge directions undetermined in the structure.  The process of variable manipulations and edge orientation updates are repeated until all edges in the current structure are oriented~\citep{cooper1999causal,eaton2007exact,he2008active,statnikov2015ultra}. Since conducting the experiments is costly, with a set of manipulated variables, the challenge of this type of methods is how to carry out a minimum number of required experiments. 

Finally, the third group of methods learns an entire causal structure from multiple interventional datasets.  Eberhardt et. al.~\citep{eberhardt2006n}  theoretically  analyzed the problem of  the constraint-based structure learning using multiple interventional datasets. Hauser and  B{\"u}hlmann~\citep{hauser2012characterization} analyzed the graph representation and greedy learning of  interventional Markov equivalence classes of DAGs. Triantafillou and Tsamardinos~\citep{triantafillou2015constraint} proposed the COmbINE algorithm to learn a joint MAG from multiple interventional datasets over overlapping variables.
Recently, He and Geng~\citep{he2016causal} proposed an algorithm to learn an entire DAG from multiple interventional datasets with unknown manipulated variables.

All these three groups of algorithms, however, were designed for finding an entire causal structure in multiple datasets, so they can be computational intractable with high-dimensional data. 

In a recent research, Peters et. al.~\citep{peters2016causal} examined the invariant property of a target variable's direct causes across different interventional datasets  and proposed the ICP algorithms. 
To discover the directed causes of a target from multiple datasets, ICP exploits the causal invariance, i.e., the conditional distribution of the target given its direct causes will remain invariant across different interventional datasets if the target is not manipulated. However, the work is for finding causes only and it is based on structural equation models~\citep{pearl2009causality}.

To summarize, existing methods for MB discovery only focus on observational data and thus are incapable of determining the structure/directions of the causal relationships. There are some methods which utilize multiple interventional datasets, but they are  either for finding the entire causal structure, or for discovering the causes of a target only. Therefore, there is a need for the solutions to specifically discovering MBs and their structures from multiple interventional datasets. 

\section{Notations and Problem Definition}\label{sec3}

In this section, we will introduce some basic definitions and mathematical notations frequently used in this paper (See Table~\ref{tb0} for a summary of the notations).
\begin{table}[t]
\centering
\caption{A summary of notations}
\begin{tabular}{|l|l|}
 \hline
\textbf{Notation}	&\textbf{Meaning}\\\hline
$DAG$	&directed acyclic graph\\\hline
$V$	& a set of random variables (vertices)\\\hline
$E$ &the edge set in a DAG\\\hline
$P$	&joint probability distribution over $V$\\\hline
$G$ &a DAG \\\hline
$M$& number of variables in $V$\\\hline
$V_i$, $V_j$, $V_k$ & a single variable in $V$\\\hline
$X$, $Y$ & a single variable in $V$\\\hline
$T$	& a given target variable in $V$\\\hline
$Z$, $S$& a subset of V, used as a conditioning set\\\hline
$V_i\indep V_j|Z$ &$V_i$ and $V_j$ are independent given $Z$\\\hline
$V_i\nindep V_j|Z$ &$V_i$ and $V_j$ are dependent given $Z$\\\hline
$sepset(V_i)$ &the conditioning set that makes $T$ and $V_i$ conditionally independent\\\hline
$\Upsilon_{i}$ &the  set of variables manipulated in the $ith$  intervention experiment \\\hline
$\Upsilon$ & \begin{tabular}{@{}l@{}}a set containing the variable sets manipulated in $n$ intervention\\ experiments respectively, i.e. $\Upsilon=\{\Upsilon_1, \cdots,\Upsilon_n\}$\end{tabular}\\\hline
$\zeta_T$ & the number of times $T$ is manipulated in the $n$ intervention experiments\\\hline
$D$ & the set comprising all the interventional datasets\\\hline
$D_i$ &\begin{tabular}{@{}l@{}}the dataset of the $ith$ intervention experiment, in which variables\\ in  $\Upsilon_i$ are intervened\end{tabular}\\\hline
$pa(T)$& the set of true parents of $T$\\\hline
$ch(T)$& the set of true children of $T$\\\hline
$pc(T)$&the true parent and children set of $T$, i.e. $pc(T)=\{pa(T)\cup ch(T)\}$
\\\hline
$sp(T)$& the set of true spouses of $T$\\\hline
$MB(T)$& the true  MB of $T$\\\hline
$MB_i(T)$ & the  MB of $T$ found in $D_i$\\\hline
$cmb_i(T)$ & the candidate MB of $T$ in $D_i$\\\hline
 $cpc(T)$ & the candidate parents and children of $T$\\\hline
$|.|$ & e.g. $|S|$, the size of the  set $S$\\\hline
$\alpha$ & significance level for independence tests\\\hline
\end{tabular}
\label{tb0}
\end{table}

Let $P$ be the joint probability distribution represented by a DAG $G$  over  a set of random variables
 $V=\{V_1,\cdots, V_M\}$.
We use $V_i\indep V_j|S$ to denote that $V_i$  and $V_j$ are conditionally independent given
$S\subseteq V\setminus\{V_i, V_j\}$, and $V_i\nindep V_j|S$ to represent that $V_i$  and $V_j$ are conditionally dependent given $S$.  The symbols $pa(V_i)$, $ch(V_i)$, and $sp(V_i)$ denote the sets of parents, children, and spouses of $V_i$, respectively.
We call the triplet $\langle V, G, P\rangle$ a \emph{Bayesian network} if $\langle V, G, P\rangle$ satisfies the Markov condition: every variable is independent of any subset of its non-descendant variables given its parents in $G$~\citep{pearl2009causality}. In a Bayesian network $\langle V, G, P\rangle$, by the Markov condition, the joint probability $P$ can be decomposed into the product of conditional probabilities as:
\begin{equation}\label{eq3-1}
\small
P(V_1, V_2,\cdots,V_M) = \prod_{i=1}^M{P(V_i|pa(V_i))}.
\end{equation}

In this paper, we consider a \emph{causal Bayesian network}, a Bayesian network in which an edge $X\rightarrow Y$ indicates that $X$ is a direct cause of $Y$~\citep{pearl2009causality,spirtes2000causation}.  For simple presentation, however, we use the term Bayesian network instead of causal Bayesian network. In the following, we present some definitions related to Bayesian networks and Markov blankets.

\begin{definition}[d-separation]~\citep{pearl2009causality} In a DAG $G$, a path $\pi$ is said to be d-separated (or blocked) by a set of vertices $S\subset V$ if and only if: (1)
$\pi$ contains a chain $V_i\rightarrow V_k\rightarrow V_j$ or a fork $V_i\leftarrow V_k\rightarrow V_j$ such that the middle vertex $V_k$ is in $S$, or
(2) $\pi$ contains an inverted fork (or collider) $V_i\rightarrow V_k\leftarrow V_j$ such that the middle vertex $V_k$ is not in $S$ and such that no descendant of $V_k$ is in $S$.
A set $S$ is said to d-separate $V_i$ from $V_j$ if and only if $S$ blocks every path from  $V_i$ to   $V_j$.
\label{def3-0}
\boxend
\end{definition}

\begin{definition}
[Faithfulness and causal sufficiency]~\citep{spirtes2000causation} Given a Bayesian network $\langle V,G,P\rangle$,  $P$ is faithful to $G$ if $\forall V_i,\ V_j\in V$, $\exists S\subseteq V\setminus\{V_i, V_j\}$ d-separates  $V_i$ and $V_j$ in $G$ if  $V_i\indep V_j|S$ holds in $P$.  Causal sufficiency denotes that any common cause of two or more variables in $V$ is also in $V$.
\boxend
\label{def3-2}
\end{definition}

\begin{theorem}\citep{spirtes2000causation} Under the faithfulness condition, given a Bayesian networks $<V,G,P>$, d-separation captures all conditional dependence and independence relations that are encoded in $G$, which implies that two variables $V_i\in V$ and $V_j\in V$ are d-separated with each other given a subset $S\subseteq V\backslash\{V_i, V_j\}$, if and only if $V_i$ and $V_j$ are conditionally independent conditioned on $S$ in $P$.
\label{the3-0}
\boxend
\end{theorem}

Theorem~\ref{the3-0} shows that conditional independence and d-separation are equivalent  if a dataset and its underlying Bayesian network are faithful to each other~\citep{spirtes2000causation}.

\begin{lemma}~\citep{spirtes2000causation}
Assuming $P$ is faithful to $G$, for $X\in V$ and $T\in V$, there is an edge between $X$ and $T$ if and only if $X\nindep T|S$, for all $S\subseteq V\backslash\{X, T\}$.
\boxend
\label{lem4-11}
\end{lemma}

Lemma~\ref{lem4-11} illustrates that if $X$ is a parent or a child of $T$, $X$ and $T$ are conditionally dependent given $\forall S\subseteq V\backslash\{X, T\}$.

Suppose that in an intervention experiment, some variables in $V$ may be manipulated.  To represent the interventions, Pearl~\citep{pearl1995causal} proposed a mathematical operator called $do(X=x)$ to indicate that $X's$ value is set to a constant $x$ by the intervention. 

If we use a DAG to represent the causal relations between variables in $V$, an intervention on a variable can be indicated by deleting all the arrows pointing to the variable~\citep{pearl2009causality}.
Let $\Upsilon_{i}\ (\Upsilon_{i}\subset V)$ be the set of variables manipulated in the $ith$  intervention experiment, $\Upsilon=\{\Upsilon_{1},\cdots,\Upsilon_{n}\}$ represent $n$ intervention experiments, and $D=\{D_1, \cdots, D_n\}$ be the $n$ corresponding interventional datasets ($D_i$ is an observational dataset if $\Upsilon_{i}=\emptyset$).  
The DAG after an intervention experiment can be defined as follows.

\begin{definition} [Post-intervention DAG]\citep{pearl2009causality}
Let $G=(V, E)$ be a DAG with variable set $V$ and edge set $E$. After the intervention on the set of variables $\Upsilon_{i}\subseteq V$ (represented as $do(\Upsilon_{i})$), the post-intervention DAG of $G$ is
$G_i=(V, E_i)$ where $E_i=\{(a,b)|(a,b)\in E, b\notin \Upsilon_{i}\}$.
The joint distribution of the post-intervention DAG $G_i$ with respect to $\Upsilon_{i}$ can be written as
\begin{equation}\label{eq3-2}
\small
P_i(V|do(\Upsilon_{i}))=\prod_{V_j\in V\setminus\Upsilon_{i}}P(V_j|pa(V_j))\times\prod_{V_j\in\Upsilon_{i}}P_i(V_j)
\end{equation}
where $P(V_j|pa(V_j))$ is the same as the conditional probability of $V_j$ in Eq.(\ref{eq3-1}) and $P_i(V_j)$ is the post-intervention conditional probability of $V_j$ after $V_j$ is manipulated.
\label{def3-4}
\boxend
\end{definition}

\begin{definition}
[Conservative rule]\citep{hauser2012characterization} If $\forall V_j\in\bigcup_{i=1}^{n}\Upsilon_i$, $\exists\Upsilon_i\in\Upsilon$ such that $V_j\notin\Upsilon_i$, then $\Upsilon$ is conservative.
\boxend
\label{def4-1}
\end{definition}

Definition~\ref{def4-1} states that given the set of $n$ intervention experiments, if for any variable that is manipulated, we can always find an experiment in which the variable is not manipulated, then we say that the set of intervention experiments is conservative.

\begin{definition}
[Markov blanket]\citep{pearl2009causality}
Under the faithfulness assumption, the Markov blanket of a target variable $T$ in a DAG, noted as $MB(T)$, is unique and consists of the parents, children and spouses of $T$, that is, $MB(T)=\{pa(T)\cup ch(T)\cup sp(T)\}$.
\label{def3-3}
\boxend
\end{definition}

Now we can define the problem to be solved in this paper as follows.

\textbf{Problem Definition:}
Given a target variable $T\in V$, this paper is focused on mining $MB(T)$ in $n$ multiple interventional datasets $D=\{D_1, \cdots, D_n\}$ without knowing $\Upsilon$. Specifically, the two tasks of the paper are defined as: 
\begin{itemize}
\item Task 1: identifying the intervention settings under which $MB(T)$ can be discovered from $D$ and $pa(T)$ can be detected.
\item  Task 2: based on the findings of task 1, developing a correct and efficient algorithm for finding $MB(T)$ and $pa(T)$ from $D$.
\end{itemize}

\section{Can we find the true MB from multiple interventional datasets?}\label{sec4}

Let $MB_i(T)$ be the MB of $T$ found in $D_i\in D$. 
In order to find the true MB of $T$ (i.e. $MB(T)$) from $D$, intuitively, the union and the intersection of the MBs discovered from all the datasets, i.e. $\bigcup_{i=1}^{n}MB_i(T)$ and $\bigcap_{i=1}^{n}MB_i(T)$, should be of interest to investigate, as the union may provide us the most information about $MB(T)$ in all datasets in $D$, whereas the intersection indicates the MB information shared by those datasets. 

In the following subsections, we will show that under different situations of manipulations, such as when $\Upsilon$ is conservative or not, the union and the intersection are closely related to $MB(T)$ or its subsets, e.g. $pa(T)$.

In the remaining sections, all lemmas and theorems are discussed under the two assumptions: (1) faithfulness and causal sufficiency, and (2) reliable independence tests.

\subsection{$T$ is not manipulated, the $\zeta_T=0$ case}

Let  $\zeta_T$ be the number of datasets in which $T$ is manipulated, and $\zeta_T=0$ represent the case that  $T$ is not manipulated in any of the $n$ datasets, i.e., for $\forall i$, $T\notin\Upsilon_i$.
In the following, we analyze the union $\bigcup_{i=1}^{n}MB_i(T)$ and the intersection $\bigcap_{i=1}^{n}MB_i(T)$ when $\zeta_T=0$  for the situation when $\Upsilon$ is conservative and not conservative, respectively.

\begin{theorem} If $\zeta_T=0$ and $\Upsilon$ is conservative, the union $\bigcup_{i=1}^{n}MB_i(T)$ is the true MB of T, i.e., $\bigcup_{i=1}^{n}MB_i(T)=MB(T)$.

\textbf{Proof:} Since $T$ is not manipulated,  then (1) By Definition~\ref{def3-4}, for $\forall D_i\in D$, the edges between $T$ and its parents are not deleted, thus by Lemma~\ref{lem4-11}, for $\forall MB_i(T)$, $pa(T)\subset MB_i(T)$ holds; (2) If $\forall V_j\in ch(T)$ and $V_j\in\bigcap_{i=1}^{n}\Upsilon_i$, by the conservative rule (Definition~\ref{def4-1}), there must exist a set $\Upsilon_k$ and $V_j\notin\Upsilon_k$. Then in $D_k$, $V_j$ is not manipulated, and the edge between $T$ and $V_j$ is not deleted. By Definition~\ref{def3-4}, $V_j\in MB_k(T)$. Since $V_j$ is not manipulated in $D_k$,  the edges between $V_j$ and its parents ($T$ and $T's$ spouses w.r.t $V_j$) are not deleted. Then the subset of $T$'s spouses w.r.t, $sp_j(T)\subseteq MB_k(T)$; (3) If $\forall V_j\in ch(T)$ and $V_j\notin\Upsilon$, $V_j$ is not manipulated. Thus, for $\forall D_i\in D$, $V_j \in MB_i (T)$, $\forall i\in\{1, \ldots, n\}$, similarly as the proof in (2), $V_j$ and the corresponding $sp(T)$ are in $MB_i(T)$.
\boxend
\label{the4-11}
\end{theorem}

\begin{theorem} If $\zeta_T=0$ and $\Upsilon$ is not conservative, $pa(T)\subseteq\bigcup_{i=1}^{n}MB_i(T)\subseteq MB(T)$.

\textbf{Proof:} (1) $T$ is not manipulated, then for $\forall MB_i(T)$, $pa(T)\subseteq MB_i(T)$ holds. (2) Since $\Upsilon$ is not conservative, if $\exists V_j\in ch(T)$ such that for $\forall\Upsilon_i\in\Upsilon$, $V_j\in\Upsilon_i$ holds, then $\forall D_i\in D$, $V_j$ is manipulated. Thus $V_j$ and the corresponding subset of $T$'s spouses $sp_j(T)$ are not in $MB_i(T)$. Then $\bigcup_{i=1}^{n}MB_i(T)\subset MB(T)$ holds. Otherwise, 
$\forall V_j\in ch(T)$ and $\exists\Upsilon_k$ such that $V_j\notin\Upsilon_k$. Then since $V_j$ is not manipulated in $D_k$, similar to the reasoning in (2) of the proof of Theorem~\ref{the4-11}, $V_j$ and $sp_j(T)$ are in $MB_k(T)$. In this case (also considering (1) which shows $pa(T)$ can be found $\forall D_i$), $\bigcup_{i=1}^{n}MB_i(T)=MB(T)$ holds.
\boxend
\label{the4-13}
\end{theorem}

Next we discuss under what condition we can identify the set $pa(T)$ (causes) via discovering $MB(T)$.

\begin{theorem} In the case of $\zeta_T=0$, if $ch(T)\subseteq\bigcup_{i=1}^{n}\Upsilon_i$, the intersection $\bigcap_{i=1}^{n}MB_i(T)=pa(T)$ holds.

\textbf{Proof:} (1) By the proofs of Theorems~\ref{the4-11} and~\ref{the4-13}, for $\forall D_i\in D$, $pa(T)\subseteq MB_i(T)$ regardless of whether $\Upsilon$ is conservative or not.
(2) Regardless of whether $\Upsilon$ is conservative or not,  if $ch(T)\subseteq\bigcup_{i=1}^{n}\Upsilon_i$, for $\forall V_j\in ch(T)$, there exists at least a set $\Upsilon_k\in\Upsilon$ and $V_j\in\Upsilon_k$, i.e., $V_j$ is manipulated in $D_k$, and hence $MB_k(T)$ does not include $V_j$ and the spouses $sp_j(T)$ w.r.t. $V_j$, so $\{\{V_j\}\cup sp_j(T)\}\nsubseteq MB_k(T)$. Therefore  from (1), $\bigcap_{i=1}^{n}MB_i(T)=pa(T)$ holds.
\boxend
\label{the4-14}
\end{theorem}

As an illustration of Theorems~\ref{the4-11} and~\ref{the4-13} , Figure~\ref{fig4-11} (a) shows the true MB of $T$ and its structure, and Figures~\ref{fig4-11} (c) to (d) are the post-intervention DAGs corresponding to three interventional datasets (green nodes are manipulated variables). From the three datasets, we obtain respectively that $MB_1=\{A\}$, $MB_2=\{A, B, C\}$, and $MB_3=\{A, B, C\}$. Thus, $\bigcup_{i=1}^{3}MB_i=\{A, B, C\}$, i.e., the MB of $T$, and $\bigcap_{i=1}^{3}MB_i=\{A\}$, that is, the parent of $T$.

\begin{figure}[t]
\centering
\includegraphics[height=1.5in,width=3in]{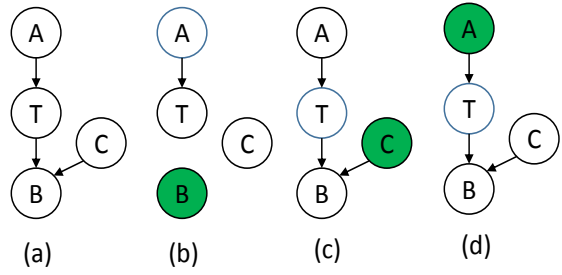}
\caption{(a) is the original DAG, while (b) to (d) are the post-intervention DAGs.}
\label{fig4-11}
\end{figure}

By Theorems~\ref{the4-11} to~\ref{the4-14}, if $\zeta_T=0$ and $ch(T)\subset\Upsilon$ hold, we achieve that (1) when $\Upsilon$ is conservative, we can get $MB(T)$ and $pa(T)$ simultaneously; (2) when $\Upsilon$ is not conservative, we may not get $MB(T)$ but are able to achieve $pa(T)$.

\subsection{$T$ is manipulated, the $0<\zeta_T<n$ case}

In this subsection, we examine the union $\bigcup_{i=1}^{n}MB_i(T)$ and the intersection  $\bigcap_{i=1}^{n}MB_i(T)$ when $T$ is intervened, for less than $n$ times, i.e. $0<\zeta_T<n$.

\begin{theorem}
If $0<\zeta_T<n$ and $\Upsilon$ is conservative, then the union $\bigcup_{i=1}^{n}MB_i(T)=MB(T)$ holds.

\textbf{Proof:} (1) As $0<\zeta_T<n$ holds, $\exists D_k\in D$, $T$ is not manipulated in $D_k$, and thus $pa(T)\subseteq MB_k(T)$ holds. (2) $\forall V_j\in ch(T)$, if $V_j$ is manipulated, since $\Upsilon$ is conservative, $\exists D_i\in D$, $V_j$ is not manipulated in $D_i$. Then $\{\{V_j\}\cup sp_j(T)\}\subseteq MB_i(T)$. Thus, $\bigcup_{i=1}^{n}MB_i(T)=MB(T)$ holds.
\boxend
\label{the4-21}
\end{theorem}

Figure~\ref{fig4-21} shows an example of applying Theorem~\ref{the4-21}. Figure~\ref{fig4-21} (a) presents the true MB of $T$ and its structure. Figures~\ref{fig4-21} (b) to (d) are the post-intervention DAGs corresponding to the three interventional datasets, indicating that $T$ is manipulated once (in Figure~\ref{fig4-21} (d)). Since $n=3$ and $\zeta_T=1$ in this example, based on Theorem~\ref{the4-21}, we have $MB(T)=MB_1(T)\cup MB_2(T)\cup MB_3(T)=\{\{A\}\cup \{A, B, C\}\cup \{B,C\}\}=\{A,B,C\}$, which is indeed the same as the true $MB(T)$ shown in Figure~\ref{fig4-21} (a).

\begin{figure}[t]
\centering
\includegraphics[height=1.5in,width=3in]{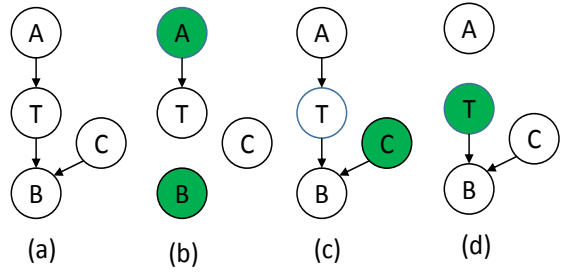}
\caption{(a) is the original DAG, while (b) to (d) are the post-intervention DAGs.}
\label{fig4-21}
\end{figure}

\begin{theorem}
If $0<\zeta_T<n$ and $\bigcup_{i=1}^{n}\Upsilon_i\setminus\{T\}$ is not conservative, $pa(T)\subseteq\bigcup_{i=1}^{n}MB_i(T)\subseteq MB(T)$.

\textbf{Proof:} (1) By the proof of Theorem~\ref{the4-21}, $pa(T)\subseteq MB_i(T)$, and thus $pa(T)\subseteq\bigcup_{i=1}^{n} MB_i(T)$, if $0<\zeta_T<n$.

(2) When $\bigcup_{i=1}^{n}\Upsilon_i\setminus\{T\}$ is not conservative, (2a) if $\forall V_j\in ch(T)$, $V_j$ is manipulated in every dataset in $D$, $\{\{V_j\}\cup sp_j(T)\}\nsubseteq MB_i(T)$ for $\forall i\in\{1,\cdots,n\}$. In the case, $pa(T)=MB_i(T)$;
(2b) if $\exists V_j\in ch(T)$ and for $\forall\Upsilon_i\in\Upsilon$, $V_j\in\Upsilon_i$ holds, $\{\{V_j\}\cup sp_j(T)\}\nsubseteq MB_i(T)$ for $\forall i\in\{1,\cdots,n\}$. Thus, $\bigcup_{i=1}^{n}MB_i(T)\subset MB(T)$; (2c) if $\forall V_j\in ch(T)$ and $\forall D_i$, $V_j$ is never manipulated, then $\bigcup_{i=1}^{n}MB_i(T)=MB(T)$.
\boxend
\label{the4-22}
\end{theorem}

\begin{figure}[t]
\centering
\includegraphics[height=1.5in,width=3in]{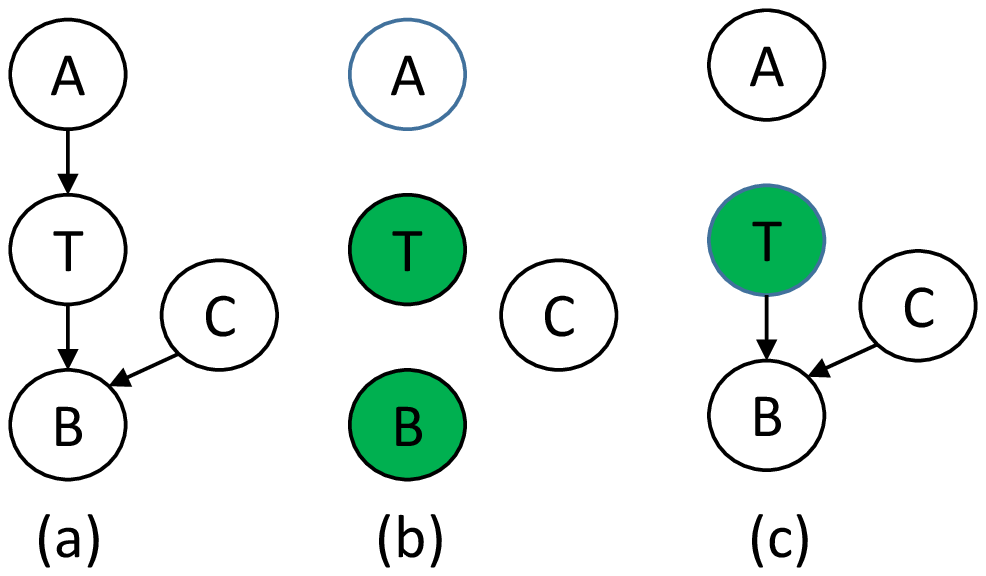}
\caption{(a) is the original DAG, while (b) to (c) are the post-intervention DAGs.}
\label{fig4-22}
\end{figure}

\begin{theorem}
In the case of $0<\zeta_T<n$, (1) if $\forall V_j\in ch(T)$, $V_j$ is manipulated, $\bigcap_{i=1}^{n}MB_i(T)=\emptyset$; (2) if $\forall\Upsilon_i$, $ch(T)\cap\Upsilon_i=\emptyset$, $\bigcap_{i=1}^{n}MB_i(T)=\{ch(T)\cup sp(T)\}$.

\textbf{Proof:} (1) (a) Since $0<\zeta_T<n$, $\exists\Upsilon_i$ such that $T\in\Upsilon_i$, and thus $pa(T)\nsubseteq MB_i(T)$ in $D_i$. Thus $pa(T)\nsubseteq\bigcap_{i=1}^{n}MB_i(T)$.
(b) Regardless of whether $\Upsilon$ is conservative or not,  if $\forall V_j\in ch(T)$, $V_j$ is manipulated, then $\forall V_j\in ch(T)$, $\exists \Upsilon_i$ and $V_j\in\Upsilon_i$ hold. This leads to $\{\{V_j\}\cup sp_j(T)\}\nsubseteq MB_i(T)$. Therefore, based on (a) and (b),
regardless of whether $\Upsilon$ is conservative or not, once $0<\zeta_T<n$ and $ch(T)\subset\bigcup_{i=1}^{n}\Upsilon_i$ hold,  $\bigcap_{i=1}^{n}MB_i(T)=\emptyset$.

(2) By the proof in (1), once $0<\zeta_T<n$ holds, $pa(T)\nsubseteq\bigcap_{i=1}^{n}MB_i(T)$.  As  $\forall\Upsilon_i$, $ch(T)\cap\Upsilon_i=\emptyset$, $\forall V_j\in ch(T)$, $V_j$ is not manipulated in any dataset. Then $\forall D_i\in D$, $MB_i(T)=\{ch(T)\cup sp(T)\}$, and $\bigcap_{i=1}^{n}MB_i(T)=\{ch(T)\cup sp(T)\}$.
\boxend
\label{the4-23}
\end{theorem}

\subsection{$T$ is manipulated, the $\zeta_T=n$ case}

As $\zeta_T=n$ holds, this means that $T$ is manipulated in each of the $n$ experiments. In this case, the following conclusions hold.

\begin{theorem}
If $\zeta_T=n$ and $\bigcup_{i=1}^{n}\Upsilon_i\setminus\{T\}$ is conservative, $\bigcup_{i=1}^{n}MB_i(T)=\{ch(T)\cup sp(T)\}$.

\textbf{Proof:} (1) If $\zeta_T=n$, then $T$ is manipulated in each dataset. Thus $\forall MB_i(T), pa(T)\nsubseteq MB_i(T)$ holds. 
(2) Since $\bigcup_{i=1}^{n}\Upsilon_i\setminus\{T\}$ is conservative, If $\exists V_j\in ch(T)$ such that $V_j$ is manipulated, there must exist a set $\Upsilon_i\in \Upsilon$ such that $V_j\notin\Upsilon_i$. Then $\{\{V_j\}\cup sp_j(T)\}\nsubseteq MB_i(T)$ holds. According to (1) and (2), $\bigcup_{i=1}^{n}MB_i(T)=\{ch(T)\cup sp(T)\}$ holds.
\boxend
\label{the4-31}
\end{theorem}

For example, Figures~\ref{fig4-22} (b) to (c) show the post-intervention DAGs corresponding to two interventional datasets, and we see that $\Upsilon_1=\{B,T\}$, and $\Upsilon_2=\{T\}$, respectively.  So $T$ is manipulated in both datasets, but without considering $T$, $\Upsilon$ is still conservative as $A$, $B$, and $C$ each are not manipulated in both datasets. Therefore, based on Theorem~\ref{the4-31}, we have $MB(T)= MB_1(T)\cup MB_2(T)=\{B, C\}$. Comparing to the true MB in Figure~\ref{fig4-22} (a), under this intervention setting, the union of the MBs found in the two datasets has missed $T$'s parent $A$.

\begin{theorem}
If $\zeta_T=n$ and $\bigcup_{i=1}^{n}\Upsilon_i\setminus\{T\}$ is not conservative, $\bigcup_{i=1}^{n}MB_i(T)\subseteq\{ch(T)\cup sp(T)\}$.

\textbf{Proof:} By the proof of Theorem~\ref{the4-31}, if $\zeta_T=n$, then for $\forall MB_i(T)$, $pa(T)\nsubseteq MB_i(T)$. When$\bigcup_{i=1}^{n}\Upsilon_i\setminus\{T\}$ is not conservative, (1) if $\exists V_j\in ch(T)$ and $\forall\Upsilon_i\in\Upsilon, V_j\in\Upsilon_i$, then for $\forall MB_i(T)$, $\{\{V_j\}\cup sp_j(T)\}\nsubseteq MB_i(T)$ holds. In this case (also noting that $pa(T)\nsubseteq MB_i(T)$),  $\bigcup_{i=1}^{n}MB_i(T)\subset\{ch(T)\cup sp(T)\}$. (2) if $\forall V_j\in ch(T)$ and $\forall\Upsilon_i\in\Upsilon, V_j\in\Upsilon_i$, the variables in $ch(T)$ are manipulated in each dataset. In this case, $\forall i, MB_i(T)=\emptyset$.
\boxend
\label{the4-32}
\end{theorem}

\begin{theorem}
If $\zeta_T=n$, (1) if $ch(T)\subseteq\bigcup_{i=1}^{n}\Upsilon_i$, $\bigcap_{i=1}^{n}MB_i(T)=\emptyset$; (2) if $\forall\Upsilon_i$, $ch(T)\cap\Upsilon_i=\emptyset$, $\bigcap_{i=1}^{n}MB_i(T)=\{ch(T)\cup sp(T)\}$.

\textbf{Proof:} (1) As $\zeta_T=n$ holds, $pa(T)\nsubseteq\bigcap_{i=1}^{n}MB_i(T)$. Following the proofs of Theorems~\ref{the4-31} and~\ref{the4-32}, regardless of whether  $\bigcup_{i=1}^{n}\Upsilon_i\setminus\{T\}$ is conservative or not, if $ch(T)\subseteq\bigcup_{i=1}^{n}\Upsilon_i$, for $\forall V_j\in ch(T)$, $\exists MB_i(T)$ such that $\{\{V_j\}\cup sp_j(T)\}\nsubseteq MB_i(T)$. Thus,  $\bigcap_{i=1}^{n}MB_i(T)=\emptyset$ holds.
(2) By the proof of Theorem~\ref{the4-23}(2), if $\forall\Upsilon_i$, $ch(T)\cap\Upsilon_i=\emptyset$, $\bigcap_{i=1}^{n}MB_i(T)=\{ch(T)\cup sp(T)\}$ holds.
\boxend
\label{the4-33}
\end{theorem}

Theorems~\ref{the4-31} to~\ref{the4-33} show that once $\zeta_T=n$, it is not possible to obtain the true $MB(T)$ via the union of the MBs found from all the datasets, as in this case, the MB discovered from each dataset does not cover $pa(T)$.

\subsection{Discussion}

\begin{table}[t]
\small
\centering
\caption{The result of $\bigcup_{i=1}^{n}MB_i$}
\begin{tabular}{|l|l|}
\hline
\multicolumn{2}{|c|}{$\zeta_T=0$} \\ \hline
$\Upsilon_{con}$        &$MB(T)$ (Theorem~\ref{the4-11})     \\ \hline
$\Upsilon_{\overline{con}}$   &$pa(T)\subseteq\bigcup_{i=1}^{n}MB_i(T)\subseteq MB(T)$ (Theorem~\ref{the4-13})    \\ \hline
\multicolumn{2}{|c|}{$0<\zeta_T<n$} \\ \hline
$\Upsilon_{con}$   &$MB(T)$ (Theorem~\ref{the4-21})    \\ \hline
$(\Upsilon\setminus T)_{\overline{con}}$    & $pa(T)\subseteq\bigcup_{i=1}^{n}MB_i(T)\subseteq MB(T)$ (Theorem~\ref{the4-22})  \\ \hline
\multicolumn{2}{|c|}{$\zeta_T=n$ } \\ \hline
$(\Upsilon\setminus T)_{con}$        &$\{ch(T)\cup sp(T)\}$ (Theorem~\ref{the4-31}) \\ \hline
$(\Upsilon\setminus T)_{\overline{con}}$    &$\bigcup_{i=1}^{n}MB_i(T)\subseteq \{ch(T)\cup sp(T)\}$  (Theorem~\ref{the4-32})  \\ \hline
\end{tabular}
\label{tb4-41}
\end{table}

\begin{table}[t]
\small
\centering
\caption{The result of $\bigcap_{i=1}^{n}MB_i(T)$}

\begin{tabular}{|l|l|}
\hline
\multicolumn{2}{|c|}{$\zeta_T=0$} \\ \hline
$ch(T)\subset\bigcup_{i=1}^{n}\Upsilon_i$        &$pa(T)$ (Theorem~\ref{the4-14})    \\ \hline
$ch(T)\nsubseteq\bigcup_{i=1}^{n}\Upsilon_i$   &$MB(T)$ (Theorems~\ref{the4-11} and~\ref{the4-13})   \\ \hline
\multicolumn{2}{|c|}{$0<\zeta_T<n$} \\ \hline
$ch(T)\subset\bigcup_{i=1}^{n}\Upsilon_i$   &$\emptyset$ (Theorem~\ref{the4-23})     \\ \hline
$ch(T)\nsubseteq\bigcup_{i=1}^{n}\Upsilon_i$    & $\{ch(T)\cup sp(T)\}$ (Theorem~\ref{the4-23}) \\ \hline
\multicolumn{2}{|c|}{$\zeta_T=n$ } \\ \hline
$ch(T)\subset\bigcup_{i=1}^{n}\Upsilon_i$        &$\emptyset$ (Theorem~\ref{the4-33}) \\ \hline
$ch(T)\nsubseteq\bigcup_{i=1}^{n}\Upsilon_i$    &$\{ch(T)\cup sp(T)\}$ (Theorem~\ref{the4-33})  \\ \hline
\end{tabular}
\label{tb4-42}
\end{table}

Tables~\ref{tb4-41} summarizes the results in Sections 4.1 to 4.3 w.r.t. the union, 
$\bigcup_{i=1}^{n}MB_i(T)$, specifically the conditions under which we can or cannot get $MB(T)$ from the union.

In Table~\ref{tb4-41},   $\Upsilon_{con}$ (or $(\Upsilon\setminus T)_{con}$)  represents that  $\Upsilon$ (or $\Upsilon\setminus T$) is conservative and $\Upsilon_{\overline{con}}$ (or $(\Upsilon\setminus T)_{\overline{con}}$) represents that $\Upsilon$ (or $\Upsilon\setminus T)$ is not conservative. From the table, we have the following observations:
\begin{itemize}
\item $\Upsilon\ or\ \Upsilon\setminus T$ is conservative: (i) If $0\leq \zeta_T<n$,  $\bigcup_{i=1}^{n}MB_i(T)=MB(T)$; (ii) If $\zeta_T=n$,  $\bigcup_{i=1}^{n}MB_i(T)=\{ch(T)\cup sp(T)\}$.

\item $\Upsilon\ or\ \Upsilon\setminus T$ is not conservative: (i) If $0\leq\zeta_T<n$, $\bigcup_{i=1}^{n}MB_i(T)$ may or may not give us $MB(T)$, but  $\bigcup_{i=1}^{n}MB_i(T)$ at least includes $pa(T)$. (ii) If $\zeta_T=n$, $\bigcup_{i=1}^{n}MB_i(T)\subseteq\{ch(T)\cup sp(T)\}$. In the worst case, $\bigcup_{i=1}^{n}MB_i(T)=\emptyset$.
\end{itemize}

Table~\ref{tb4-42} summarizes under what manipulation settings we are able to identify the set $pa(T)$ (i.e. the causes of $T$) via the intersection of the MBs discovered from all the datasets. Here are our findings:
\begin{itemize}
\item If $\zeta_T=0$ and $ch(T)\subset\bigcup_{i=1}^{n}\Upsilon_i$, $\bigcap_{i=1}^{n}MB_i(T)=pa(T)$ regardless of whether $\Upsilon$ is conservative or not. 
As long as $T$ is not manipulated, we can identify $pa(T)$ in multiple interventional datasets. This result has high practical significance, because in practice, when experiments are conducted, we do not manipulate the target of interest. 
And this result can significantly improve the computational efficiency for revealing causes of a variable  only via discovering the MB of the variable without learning a global structure or a local structure.

\item If $0<\zeta_T\leq n$ , $pa(T)\nsubseteq\bigcap_{i=1}^{n}MB_i(T)$ regardless of whether $\Upsilon$ is conservative or not.  In this case, $\bigcap_{i=1}^{n}MB_i(T)\subseteq \{ch(T)\cup sp(T)\}$.
\end{itemize}

\section{The MIMB algorithm}\label{sec5}

In section 4, we have investigated the conditions under which the true MB and the parent set of $T$ can be found from multiple interventional datasets. In this section, we propose a new and efficient algorithm, MIMB to mine \underline{M}ultiple  \underline{I}nterventional datasets for \underline{M}arkov  \underline{B}lanket discovery, without knowing the manipulated variables.
Under the conditions identified in Section 4, MIMB can return the true MB of $T$ and the parent set of $T$. 
Before discussing the MIMB algorithm, we first present a baseline algorithm for evaluating MIMB.

\subsection{The baseline algorithm}
The baseline algorithm (Algorithm 1) finds the MB of $T$ in each dataset independently, then uses the union of those MBs as $MB(T)$ and the intersection of the MBs as $pa(T)$.

At Step 2,  the baseline algorithm employs
the  HITON-MB algorithm~\citep{aliferis2003hiton,aliferis2010local1}, a state-of-the-art MB discovery algorithm with a single observational dataset to discover the MBs from each dataset (any other sound MB discovery methods can be used here). HITON-MB contains two steps. Firstly, it finds the parents and children of a given target variable. Then based on the found parents and children, HITON-MB identifies the spouses of the target. Without learning an entire Bayesian network, HITON-MB is able to effectively and efficiently find the MB of a given target from an observational dataset containing thousands of variables.

\begin{theorem} In Algorithm 1, (1) if $0\leq\zeta_T<n$, $base\_MB(T)=MB(T)$, and (2) if $\zeta_T=0$ and  $ch(T)\subset\bigcup_{i=1}^{n}\Upsilon_i$, $base\_pa(T)=pa(T)$.

\textbf{Proof:}  By the assumptions of faithfulness and reliable tests made at the beginning of Section 4 (i.e. supposing that the set $MB_i(T)$ found in $D_i$ at Step 2 of Algorithm 1 is correct), if $\zeta_T=0$ or $0<\zeta_T<n$, by Theorems~\ref{the4-11} and~\ref{the4-21}, (1) holds. By Theorem~\ref{the4-14}, (2) holds.
\boxend
\label{the5-ext1}
\end{theorem}

Theorem~\ref{the5-ext1} illustrates that the baseline algorithm is theoretically sound. However, it is not efficient, because it has to conduct independence tests to find the MB in each dataset. But if $X\indep T|S\ (S\subseteq V\setminus \{X,T\}$ and $S\neq\emptyset)$ holds in one dataset, we may not need to conduct the test in the remaining datasets. Moreover, since Algorithm 1 simply takes the union of the MBs found in each dataset at Step 4, in practice, due to data noise or sample bias,  false positives may be introduced to the MBs found in some datasets, and the false discoveries may be carried over into the final output of the algorithm.

Thus, to overcome the issues with the baseline algorithm, in next section, we propose the MIMB algorithm to leverage the information across multiple datasets.

\begin{algorithm}[t]
\SetKwData{Left}{left}
  \SetKwData{Up}{up}
  \SetKwFunction{FindCompress}{FindCompress}

  \SetKwRepeat{Repeat}{repeat}{until}
\KwInput{$D=\{D_1,D_2,\cdots,D_n\}$, T: the target variable}

\KwOutput{base\_MB(T), base\_pa(T)}

\For {i=1 to n}
{
	Find $MB_i(T)$ in dataset $D_i$ /*HITON-MB*/
}

$base\_MB(T)=\bigcup_{i=1}^{n}MB_i(T)$

$base\_pa(T)=\bigcap_{i=1}^{n}MB_i(T)$

\caption{\textit{The Baseline Algorithm}}
\label{algorithm 1}
\end{algorithm}



\begin{algorithm}[t]
\SetKwData{Left}{left}
  \SetKwData{Up}{up}
  \SetKwFunction{FindCompress}{FindCompress}

  \SetKwRepeat{Repeat}{repeat}{until}
\KwInput{$D=\{D_1,D_2,\cdots,D_n\}$, T: the target variable}

\KwOutput{$mimb\_MB(T)$, $mimb\_pa(T)$}

 $[cpc(T), cmb, sepset]=MIPC(D,T)$;

\For {each variable $V_j\in cpc(T)$}
{ $cpc(V_j)=MIPC(D,V_j)$;/*here, MIPC only needs to output $cpc(V_j)$*/

  \For{$\forall X\in cpc(V_j)\setminus\{T\cup cpc(T)\}$}
    {
        \If{$\exists k\in\{1,\cdots,n\}, V_j\in cmb(k)$ s.t.\\
         $X\indep T|sepset(X)$ and $X\nindep T|\{sepset(X)\cup V_j\}$ in $D_k$}
           {
              $cmb_k(T)=cmb_k(T)\cup X$;
             }
}
}

$mimb\_MB(T)=\bigcup_{i=1}^{n}cmb_i(T)$

$mimb\_pa(T)=\bigcap_{i=1}^{n}cmb_i(T)$

Output $mimb\_MB(T)$ and $mimb\_pa(T)$
\caption{\textit{The MIMB Algorithm}}
\label{algorithm 2}
\end{algorithm}

\subsection{The proposed MIMB algorithm}

\subsubsection{Overview of MIMB}

Suppose $cpc(T)$ keeps the candidate parents and children of $T$, $|cpc(T)|$ means the number of variables in $cpc(T)$, and $cmb=\{cmb_1(T),\cdots,cmb_n(T)\}$ where $cmb_i(T)$ stores the candidate MB of $T$ in $D_i$. $sepset(V_j)$ is the conditioning set that may make $T$ and $V_j$ conditionally independent  and $pc(T)$ denotes the true parents and children set of T, i.e., $pc(T)=\{pa(T)\cup ch(T)\}$. For notation convinience, we use notation $sepset$ to represent conditioning sets of all non-target variable, i.e., $sepset=\{sepset(V_1),\cdots, sepset(V_j),\cdots, sepset(V_M)\}$.
MIMB (Algorithm 2) takes the union of $cmb_i(T)$, i.e, $mimb\_MB(T)$, and the intersection of $cmb_i(T)$, that is, $mimb\_pa(T)$, as its output. However, compared to the baseline algorithm, MIMB  leverages multiple datasets to improve the efficiency and reduce false positives in the output. Based on the discussion in Section 4.4,  MIMB will return the true MB of $T$ if $0\leq \zeta_T<n$, while in other cases, MIMB may not. And if $\zeta_T=0$, MIMB will return the true parents of $T$.

To deal with multiple interventional datasets,  in Algorithm 2, MIMB includes a new subroutine, MIPC to mine \underline{M}ultiple \underline{I}nterventional datasets to find \underline{P}arents and \underline{C}hildren of $T$ at Step 1 and  it discovers spouses of $T$ at Steps 2 to 10. 
We leave the discussion of the MIPC subroutine to Section 5.2.2, and in the following, we discuss the details of the strategy of identifying  spouses of $T$ from multiple interventional datasets.


 
Based on its definition, the spouse set of $T$, i.e. $sp(T)$ comprises the parents (excluding $\{T\cup pc(T)\}$) of all of $T$'s children, i.e. $sp(T)=\{\bigcup_{V_i\in ch(T)}pa(V_i)\}\setminus\{T\cup pc(T)\}$. Accordingly, Algorithm 2 (Step 3) firstly finds the parents and children of each variable $V_j$ in $cpc(T)$, using the MIPC subroutine. Since MIPC finds the parents and children of a variable without distinguishing between the parents and children, based on  Theorem~\ref{the5-10} below, in Steps 4 to 9, Algorithm 2 determines whether the parents and children of $V_j$ found in Step 3 is a true spouse of $T$ or not, while avoiding checking all datasets.
 
\begin{theorem}~\citep{spirtes2000causation} Let $V_j\in pc(T)$ and $V_k\in pc(V_j)\setminus\{T\cup pc(T)\}$, if $\exists S\subseteq V\setminus\{T,V_k,V_j\}$, both $V_k\indep T|S$ and $\ V_k\nindep T|\{S\cup V_j\}$ hold,  $V_k\in sp(T)$.
\label{the5-10}
\boxend
\end{theorem}

At Steps 4 to 9 of Algorithm 2, the MIMB algorithm employs Theorem~\ref{the5-10} and the property of manipulated variables (i.e. Definition~\ref{def3-4}) to avoid checking all datasets. 
Assuming $Y\in ch(T)$, i.e., a child of $T$, and $X$ is a spouse of $T$ through $Y$, By Definition~\ref{def3-4}, if $Y$  is manipulated in $D_i$,  $Y\notin ch(T)$ and $X\notin pa(Y)$ in $D_i$. In this case,  $X$ is not a spouse of $T$ through $Y$ in $D_i$. 
Otherwise, if $Y$ is not manipulated in $D_i$, there must exist a path from $T$ to $X$ through $Y$ to form a v structure, i.e., $T\rightarrow Y\leftarrow X$. Then if $\exists S\subseteq V\setminus\{T,X\}$ such that $T\indep X|S$ and $T\nindep X|\{S,Y\}$ hold in $D_i$, then by Theorem~\ref{the5-10},  $X$ is a spouse of $T$. 

For example, in a data set, the child of $T$, $C$ is manipulated, based on Definition~\ref{def3-4}, the post-intervention DAG will be as in Figure~\ref{fig5-new0} (b) where $T\indep  C$ and $T\indep E$.
There does not exist a path from $T$ to $E$ through $C$, i.e., the v structure $T\rightarrow C\leftarrow E$ does not exist.  Assuming $cmb_{a}(T)$ denotes the MB of $T$ found by the MIPC subroutine in Figure~\ref{fig5-new0} (a),  in Figure~\ref{fig5-new0} (a), $C$ is not  manipulated, and thus $T\nindep  C$ and $T\indep E$ (i.e. $S$ is an empty set in the example). Then in Figure~\ref{fig5-new0} (a), $C\in cmb_{a}(T)$ and $T\nindep E|C$ hold.  According to Theorem~\ref{the5-10}, $E$ is a spouse of $T$.
However, in Figure~\ref{fig5-new0} (b), $C\notin cmb_{b}(T)$ holds. In this case, $T\indep E|C$ holds, then $E$ is not  a spouse of $T$. 

\begin{figure}[t]
\centering
\includegraphics[height=1.2in,width=2.6in]{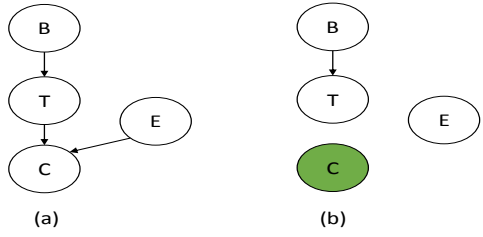}
\caption{The relation among the target $T$, the child $C$ of $T$, and the spouse $E$ of $T$ when the child $C$ is not manipulated and manipulated (manipulated nodes in green)}
\label{fig5-new0}
\end{figure}

Since we do not know which variables are manipulated and which variables are the children of $T$ in $cpc(T)$ computed by the MIPC subroutine, MIMB uses $cmb_i(T)$ to keep the candidate MB of $T$ in $D_i$. If $Y\in cmb_i(T)$ holds, this at least illustrates that in $D_i$, $Y$ is dependent on $T$, and thus $Y$ may not manipulated in $D_i$. Moreover, if  both $T\indep X|S$ and $T\nindep X|\{S,Y\}$ hold in $D_i$, based on Theorem~\ref{the5-10}, $X$ is a spouse of $T$.
We summarise the above discussion below as Corollary~\ref{cor5-1}, which is followed by steps 4 to 9 of Algorithm 2.
\begin{corollary}
Referring to Algorithm 2, let $V_j\in cpc(T)$ and $V_k\in cpc(V_j)\setminus\{T\cup cpc(T)\}$, if $\exists D_i\in D$ such that $V_j\in cmb_i(T)$ and $\exists S\subseteq V\setminus\{T,V_k,V_j\}$ both $V_k\indep T|S$ and $\ V_k\nindep T|\{S\cup V_j\}$ hold,  then $V_k\in sp(T)$.
\label{cor5-1}
\boxend
\end{corollary}

\begin{algorithm}[t]
\SetKwData{Left}{left}
  \SetKwData{Up}{up}
  \SetKwFunction{FindCompress}{FindCompress}

  \SetKwRepeat{Repeat}{repeat}{until}
\KwInput{$D=\{D_1,D_2,\cdots,D_n\}$, T: the target variable}

\KwOutput{cpc(T), cmb, sepset}
$cpc(T)=\emptyset$; $cmb_i(T)=\emptyset\ (\forall i\in\{1,\cdots,n\})$; $ipc(T)=\emptyset$;

\For {j=1 to $|V\setminus T|$}
{

\For {i=1 to n}
{
     \If {$V_j\nindep T$ in $D_i$}
{
      $cpc(T)=cpc(T)\cup V_j$;

      $cmb_i(T)=cmb_i(T)\cup V_j$;

}
}
$sepset(j)=\emptyset$;
}

\For {each variable $V_j\in cpc(T)$}
{
 \If{$\exists S\subseteq ipc(T)$$\&$$S\neq\emptyset$ s.t. $\exists k\in\{1,\cdots,n\}$, $V_j\in cmb(k)$ $\&$ $S\subset cmb(k)$ $\&$ $T\indep V_j|S$ in $D_k$}
        {
            
            $cmb_i(T)=cmb_i(T)\setminus  V_j\ (\forall i\in\{1,\cdots,n\})$;
            
  
              $sepset(V_j)=S$;

              goto 11;
             }
$ipc(T)=ipc(T)\cup V_j$;

\For {$\forall Y\in ipc(T)\setminus V_j$}
{
 \If{$\exists S\subseteq ipc(T)\setminus Y$ $\&$$S\neq\emptyset$ s.t. $\exists k\in\{1,\cdots,n\}$, $Y\in cmb(k)$ $\&$ $S\subset cmb(k)$ $\&$ $T\indep Y|S$ in $D_k$}
        {
            
             $ipc(T)=ipc(T)\setminus Y$;
             
             \If {$\exists h\in\{1,\cdots,n\}$ such that  $Y\in cmb(h)$}
               {
                   $cmb(h)=cmb(h)\setminus  Y$;
                }
             
              $sepset(Y)=S$;
             }
}
}
$cpc(T)=ipc(T)$

$cmb=\bigcup_{i=1}^ncmb_i(T)$

Output $cpc(T)$, $cmb$, and  $sepset$
\caption{\textit{The MIPC Subroutine}}
\label{algorithm 3}
\end{algorithm}

\subsubsection{The MIPC Subroutine}

The MIPC subroutine (Algorithm 3) is designed for finding parents and children of $T$. 
Steps 2 to 10, i.e., the first for loop in Algorithm 3, discover the candidate parents and children which at this stage are those variables dependent on the target $T$, and add them to $cpc(T)$ and to the candidate MB found in current datasets, i.e., $cmb=\bigcup_{i=1}^ncmb_i(T)$. For each variable $V_j\in V\setminus T$, if $V_j$ is dependent on $T$ in a dataset $D_i$, Steps 5 and 6 add it to $cpc(T)$ and $cmb_i(T)$, respectively. Otherwise, if $V_j\indep T$ holds in all $n$ datasets, Algorithm 3 never considers $V_j$ as a candidate for being added to both $cpc(T)$ and $cmb_i(T)$ again.

Both $cpc(T)$ and $cmb$ found at Steps 2 to 10 contain variables correlated to T, but we only want the parents and children of $T$ to be included in them (for $cmb$, spouses of $T$ will be added at Steps 2 to 10 in Algorithm 1), so false positives need to be removed.
False positives are those non descendants excluding parents (for example, node $A$  in Figure~\ref{fig5-new01}) and those descendants excluding children (for instance, node $D$  in Figure~\ref{fig5-new02}). They can be removed based on the Markov condition.
In a Bayesian network, the Markov condition denotes that for $\forall V_j\in V$, conditioning on its all parents, $pa(V_j)$, $V_j$ is independent of its non-descendants. However, with multiple interventional datasets, how do we use the Markov condition to remove those false positives from $cpc(T)$ and $cmb$ added at Steps 2 to 10? In the following, we will discuss under what kind of manipulations in a dataset within $D$ those non-descendants and descendants of $T$ can be removed as false positives or cannot be removed from $cpc(T)$.
\begin{itemize}
 \item For a $T$'s non-descendant $X\notin pa(T)$, $X$ is added to $cpc(T)$ at Steps 2 to 10. In an observational dataset,  by the Markov condition, conditioning on $pa(T)$, $T\indep X|pa(T)$ holds. If $T$ is manipulated in $D_i\in D$,  by the manipulation rule (Definition~\ref{def3-4}), for $\forall V_j\in pa(T)$, $V_j$ is independent of $T$ such that  $pa(T)\nsubseteq cmb_i(T)$ in $D_i$. In this case, we cannot determine whether $X$ is a $T$'s non-descendant in $D_i$.

Otherwise, if $T$ is not manipulated in $D_i$, $pa(T)\subseteq cmb_i(T)$ in $D_i$ and  there must exist a subset $S\subseteq pa(T)$ and $S\subseteq cmb_i(T)$ in $D_i$ such that $X\indep T|S$ in $D_i$.  Thus, $X$ can be removed from  $cpc(T)$ as a $T$'s non-descendant in $D_i$. Based on the observations above, we can conclude that in $D_i\in D$, if $X\in cmb_i(T)$ and $S\subseteq cmb_i(T)$, there is at least a directed path from $X$ to $T$ through $S\subseteq pa(T)$ (i.e. $T$ is not manipulated). In this case, if $X\indep T|S$ holds, $X$ as non-descendant of $T$ can be removed from $cpc(T)$.  

\textbf{Example 1.} Figure~\ref{fig5-new01} illustrates what kind of manipulations in a dataset we can use to remove a non-descendant of $T$. In Figure~\ref{fig5-new01}(a) to (b),  the manipulated variables are in blue.  In Figure~\ref{fig5-new01} (b), $T$ is manipulated, $T\indep  B$ and $T\indep A$. In this case, $\{A\cup B\}\nsubseteq cmb_{b}(T)$, and thus we cannot determine whether $A$ is a $T$'s non-descendant in Figure~\ref{fig5-new01} (b). However,  in Figure~\ref{fig5-new0} (a), $T$ is not  manipulated. Thus,  there exists a  directed path from $A$ to $T$ through $B$, by the Markov condition, that is, $\{A\cup B\}\subseteq cmb_{b}(T)$. Then $T\indep A|B$ holds and $A$ can be removed as a  $T$'s non-descendant. 

\item For a $T$'s descendant $Y\notin ch(T)$ and $Y\in cpc(T)$ after Steps 2 to 10, assuming $pa(Y)$ is the parent set of $Y$.  By the Markov condition, given $pa(Y)$, $T\indep Y|pa(Y)$ holds, and $Y$ is a $T$'s descendant.
Assuming each variable in $pa(Y)$ is manipulated.  (1) If $\{pa(Y)\cap pa(T)\}$ is empty (i.e. no variables are parents of both $T$ and $Y$), $pa(Y)\nsubseteq cmb_i(T)$, and thus, we cannot determine whether $Y$ is a $T$'s descendant in $D_i$.
(2) If $\{pa(Y)\cap ch(T)\}$ is not empty, for $\forall X\in \{pa(Y)\cap ch(T)\}$,  $X$ will be independent of $T$ in $D_i$, since $X$ as a child of $T$ is manipulated. Accordingly, $\{pa(Y)\cap ch(T)\}\nsubseteq cmb_i(T)$. (3) If $\{pa(Y)\cap pa(T)\}$ is not empty, for $\forall X\in \{pa(Y)\cap pa(T)\}$,  $X$ will be dependent on $T$ in $D_i$, since as a parent of  both $T$ and $Y$, manipulating $X$ does not cut the path on $Y$ to $T$ through $X$, Thus in this case, $\{pa(Y)\cap pa(T)\}\subseteq cmb_i(T)$.  

In summary, (1) If each variable in $pa(Y)$ is manipulated but $Y\in cmb_i(T)$ holds in $D_i$, there must exist a path on $Y$ to $T$ through $S=\{pa(Y)\cap pa(T)\}$ and $S\subseteq cmb_i(T)$. In this case, if $T\indep Y|S$,  $Y$ is a $T$'s descendant in $cpc(T)$ (See Figure~\ref{fig5-new02}(b) in Example 2). (2) If $\exists S\in\{pa(Y)\cap ch(T)\}$ is not manipulated in $D_i$ and $S\subseteq cmb_i(T)$, there must exist a path from $T$ to $Y$ through $S$ in $D_i$. If  $X\indep T|S$ holds,  and $X$ is a $T$'s descendants (See Figure~\ref{fig5-new02}(c) in Example 2).

\textbf{Example 2.} Figure~\ref{fig5-new02} illustrates what kind of datasets we can use to remove a descendant of $T$. In Figure~\ref{fig5-new02}(a) to (c), the manipulated variables are in blue.
In Figure~\ref{fig5-new02}(b), for the $T$'s descendant $D$, $pa(D)=\{F,C,E\}$, $\{pa(Y)\cap ch(T)\}=\{F,C\}$, and $\{pa(Y)\cap pa(T)\}=\{E\}$. Although the variables in $pa(D)=\{F,C,E\}$ are all manipulated, there still exists a path from $D$ to $T$ through $E$ and $\{E,D\}\subseteq cmb_{(b)}(T)$. In  Figure~\ref{fig5-new02}(b), $E\in pa(D)$ and $T\indep D|E$ holds. Thus, $D$ is a $T$'s descendant.
Figure~\ref{fig5-new02}(c) gives an example of not all variables in $\{pa(Y)\cap ch(T)\}$ being manipulated. In Figure~\ref{fig5-new02}(c), both $T$ and $F\in pa(D)$ are manipulated. Since $C\in ch(T)$ is not manipulated, there exists a path from $D$ to $T$ through $C$ and $\{C,D\}\subseteq cmb_{(c)}(T)$. In  Figure~\ref{fig5-new02}(c), $C\in pa(D)$ and $T\indep D|C$ holds. Thus, $D$ is a $T$'s descendant.

\end{itemize}

However, to remove those false positives,  the challenge is that we do not know which variables are manipulated in each dataset and for each variable in $cpc(T)$ obtained at Steps 2 to 10, we do not known which variable is a parent, or a child, or a non-descendant, or a descendant of $T$.  
From the discussions above, we can see that whatever for a $T$'s non-descendant or a $T$'s descendant, both called $Y$, if $T\indep Y|S$ holds in $D_i$ and $Y$ can be removed from $cpc(T)$, dataset $D_i$ should satisfy that $\{Y\cup S\}\subseteq cmb_i(T)$ must hold in $D_i$, i.e., there must exist a path from $Y$ to $T$ through $S$. 

We formalize the idea to identify and remove false positives in $cpc(T)$ at Steps 11 to 27 in Algorithm 3 in Corollary~\ref{cor5-2} below. If a false positive is detected in $D_i\in D$, we remove it from $cpc(T)$ and $cmb_j(T)$ (for $\forall j\in\{1,\cdots,n\}$), avoiding checking all datasets.

\begin{corollary}
Referring to Algorithm 3, assuming $V_j\in ipc(T)$ and $\exists S\subseteq \{ipc(T)\setminus V_j\}$, if $\exists k\in\{1,\cdots,n\}$ such that $\{V_j\cup S\}\subseteq cmb_k(T)$ and $V_j\indep T|S$ in $D_k$, $V_j\notin pc(T)$.
\boxend
\label{cor5-2}
\end{corollary}

\begin{figure}[t]
\centering
\includegraphics[height=1.2in,width=3in]{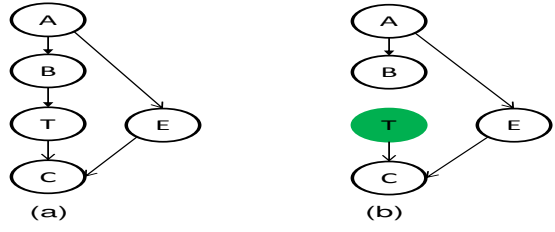}
\caption{Removing non-descendant $A$ of $T$: the target $T$, $T$'s parent $B$, and $T$'s non-descendant $A$ (manipulated nodes in green)}
\label{fig5-new01}
\end{figure}

\begin{figure}[t]
\centering
\includegraphics[height=1.3in,width=4in]{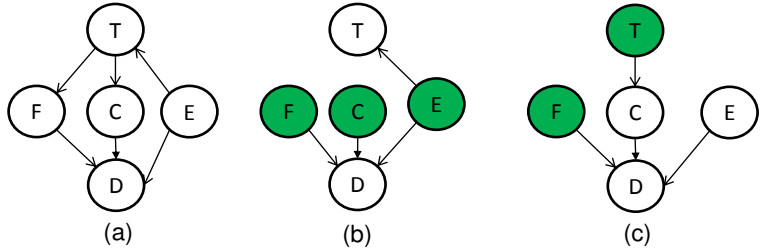}
\caption{Removing descendant $D$ of $T$: the target $T$, $T$'s child $C$, and $T$'s descendant $D$ (manipulated nodes in green)}
\label{fig5-new02}
\end{figure}

Using Corollary~\ref{cor5-2} to determine whether a variable in $cpc(T)$ is a false positive, the naive approach is to check all the subsets of $cpc(T)$ for finding a subset $S$ in a dataset satisfying the corollary.  To improve efficiency,  Algorithm 3 (Steps 11 to 27) uses an intermediate parent and child set $ipc(T)$ to greedliy sequentially update $cpc(T)$ and $cmb$ ($cmb=\bigcup_{i=1}^ncmb_i(T)$), that is, removing false positives from $cpc(T)$ and $cmb$, via conditional independence tests. Moreover,  instead of using all datasets in $D$, for $\forall V_j\in cpc(T)$, Steps 11 to 27 attempt to avoid checking all datasets in determining a subset $S$ in a dataset satisfying Corollary~\ref{cor5-2}.

At each iteration, at Steps 12 to 16, Algorithm 3 takes a variable $V_j$ from $cpc(T)$, then finds  in which dataset there exists a subset $S\subseteq ipc(T)\setminus V_j$ satisfying Corollary~\ref{cor5-2}. If finding the subset $S$ and $V_j\indep  T|S$ holds in a dataset, Algorithm 3 considers next variable in $cpc(T)$. 
If not, Step 17 adds $V_j$ to $ipc(T)$ and Steps 18 to 26 are triggered.  At Steps 18 to 26,  for each variable $Y$ currently in $ipc(T)$, due to the inclusion of $V_j$ in $ipc(T)$ just now at Step 17, if there exists $S$, a subset of $ipc(T)$ satisfying Corollary~\ref{cor5-2} in a dataset, such that $Y$ is independent of $T$ given $S$ in the dataset, $ Y$ can be removed from the current parent and children set $ipc(T)$ (Step 20) and from the current MB set in each dataset, i.e. $cmb_i(T)$ at Steps 21 and 22 if $\exists i\in\{1,\cdots,n\}$ such that $Y\in cmb_i(T)$.
The second for loop in Algorithm 3 (Steps 11 to 27) will terminate until all variables in $cpc(T)$ are checked once.  

Since the size of $ipc(T)$ grows gradually and can be kept as small as possible, this may avoid an expensive search for a subset of large size in $cpc(T)$. In addition, MIPC includes an optimization technique at Step 19 for time efficiency. Once a variable is added to $ipc(T)$ at Step 17,  at Steps 18 to 26, for each variable $Y$ in $ipc(T)$, MIPC will check the subsets within the set $\{ipc(T)\setminus Y\}$ that only include the newly added variable, instead of all subsets within $ipc(T)\setminus Y$.

Finally, Steps 28 to 30 output $cpc(T)$, $cmb$, and $sepset$. The set $sepset$ includes conditioning sets for all variables in $V$ but the target $T$. For $cpc(T)$ and $cmb$, in addition to all parents and children of $T$, they may contain some false positives. We will discuss the set $cpc(T)$ found by Algorithm 3 in Theorems~\ref{the5-23} and~\ref{the5-24} below in Section 5.3.

\begin{figure}[t]
\centering
\includegraphics[height=2.8in,width=3.9in]{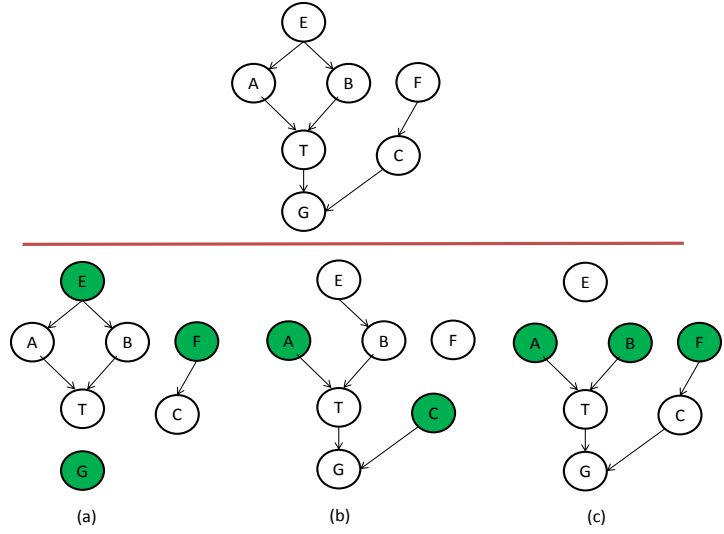}
\caption{An example of tracing the MIMB algorithm (Figures~\ref{fig5-new2} (a) to (c) are manipulated from the top DAG (manipulated nodes in green))}
\label{fig5-new2}
\end{figure}

\subsubsection{Tracing the MIMB Algorithm}
In the following we use the example in  Figure~\ref{fig5-new2} to walk through the proposed MIMB algorithm.
In Figure~\ref{fig5-new2}, the top graph is a DAG showing the true causal relationships among the set of variables, where $T$ is the target of interest. The diagrams at the bottom of Figure~\ref{fig5-new2} ((a), (b) and (c)) are the post-intervention DAGs corresponding to three interventional datasets. According to Algorithm 2, we firstly apply MIPC (Algorithm 3) to find the parent and children set of $T$, $cpc(T)$. The following shows how Algorithm 3 goes through its Steps 2 to 10 to find out $cpc(T)$ from the set of non-target variables $\{E, A, B, F, C, G\}$. 

\begin{itemize}
\item $E$: Initially, $cpc(T)=\emptyset$, $cmb_{a}(T)=\emptyset$, $cmb_{b}(T)=\emptyset$, and $cmb_{c}(T)=\emptyset$. As $E\nindep T$ in Figures~\ref{fig5-new2} (a) and (b), $cpc(T)=\{E\}$ and $cmb_{a}(T)=\{E\}$ and $cmb_{b}(T)=\{E\}$ and $cmb_{c}(T)=\emptyset$.

\item $A$: As $A\nindep T$ in Figures~\ref{fig5-new2} (a) to (c), $cpc(T)=\{E, A\}$, $cmb_{a}(T)=\{E, A\}$, $cmb_{b}(T)=\{E, A\}$, and $cmb_{c}(T)=\{A\}$.

\item $B$: As $B\nindep T$ in Figures~\ref{fig5-new2} (a) to (c), $cpc(T)=\{E, A, B\}$ and $cmb_{a}(T)=\{E, A, B\}$, $cmb_{b}(T))=\{E, A, B\}$, and $cmb_{c}(T)=\{A, B\}$.

\item $F$: As $F\indep T$ in Figures~\ref{fig5-new2} (a) to (c), $cpc(T)=\{E, A, B\}$ and $cmb_{a}(T)=\{E, A, B\}$, $cmb_{b}(T)=\{E, A, B\}$, and $cmb_{c}(T)=\{A, B\}$.

\item $C$: As $C\indep T$ in Figures~\ref{fig5-new2} (a) to (c), $cpc(T)=\{E, A, B\}$ and $cmb_{a}(T)=\{E, A, B\}$, $cmb_{b}(T)=\{E, A, B\}$, and $cmb_{c}(T)=\{A, B\}$.

\item $G$: As $G\nindep T$ in Figures~\ref{fig5-new2} (b) to (c), $cpc(T)=\{E, A, B, G\}$ and $cmb_{a}(T)=\{E, A, B\}$, $cmb_{b}(T)=\{E, A, B, G\}$, and $cmb_{c}(T)=\{A, B, G\}$.
\end{itemize}

So after carrying out Steps 2 to 10 of Algorithm 3, we get $cpc(T)=\{E, A, B, G\}$ and $cmb_{a}(T)=\{E, A, B\}$, $cmb_{b}(T)=\{E, A, B, G\}$, and $cmb_{c}(T)=\{A, B, G\}$. Then Steps 11 to 27 are implemented as follows

\begin{itemize}
\item $E$: Since  $ipc(T)$ is empty, $E\in cmb_{a}(T)$ and $E\in cmb_{b}(T)$, and $E\nindep T$ in Figures~\ref{fig5-new2} (a) and (b),  then $ipc(T)=\{E\}$ and move to next variable within $cpc(T)$.

\item $A$: Step 12 firstly examines which datasets make $A\indep T|E$ hold. Since $cmb_{a}(T)=\{E, A, B\}$ and $cmb_{b}(T)=\{E, A, B, G\}$ include $E$ and $A$, then Step 12 checks whether $A\indep T|E$ holds in Figures~\ref{fig5-new2} (a) and (b). But  $A\indep T|E$ does not hold in both figures.  Then $ipc(T)=\{E, A\}$. Next,  Step 20 examines which datasets make $E\indep T|A$ hold. Since  $E\nindep T|A$ holds in Figures~\ref{fig5-new2} (a) and (b), finally, $ipc(T)=\{E, A\}$.

\item $B$: $ipc(T)$. Step 12 examines which datasets makes $B\indep T|E$, $B\indep T|A$, and $B\indep T|\{E, A\}$ hold or not, respectively. $B\indep T|E$, $B\indep T|A$, and $B\indep T|\{E, A\}$ do not hold in Figures~\ref{fig5-new2} (a) to (c). Then $ipc(T)=\{E, A, B\}$.
Next, Step 20 checks which datasets makes  $A\indep T|B$, and $A\indep T|\{E, B\}$ hold or not, respectively. Those three terms also do not hold in Figures~\ref{fig5-new2} (a) to (c). 
Finally,  Step 20 examines which datasets makes $E\indep T|B$, and $E\indep T|\{A, B\}$ hold or not, respectively. Since $E\indep T|B$ holds in Figure~\ref{fig5-new2} (b), then Steps 22 to 23 remove $E$ from  $ipc(T)$,  $cmb_{a}(T)$, and $cmb_{b}(T)$. Finally, $ipc(T)=\{A, B\}$.

\item $G$: Step 12 examines which datasets makes $G\indep T|A$, $G\indep T|B$, and $G\indep T|\{B, A\}$ hold or not, respectively. The three terms do not hold in Figure~\ref{fig5-new2} (b) and Figure~\ref{fig5-new2}(c). Then $ipc(T)=\{A, B, G\}$. Next, Step 20 checks whether $A\indep T|G$, and $A\indep T|\{B, G\}$ hold or not and whether  $B\indep T|G$, and $B\indep T|\{A, G\}$ hold or not, respectively. Finally, $ipc(T)=\{A, B, G\}$.
\end{itemize}

Thus, after checking all the elements in $cpc (T)$, we get  $cpc(T)=\{A, B, G\}$, $cmb_{a}(T)=\{A, B\}$, $cmb_{b}(T)=\{A, B, G\}$, and $cmb_{c}(T)=\{A, B, G\}$. By Step 29, we get  $cmb=\{A, B, G\}$ and $sepset(F)=\emptyset$, $sepset(C)=\emptyset$, and $sepset(E)=\{B\}$.
Accordingly, after Step 2 in Algorithm 2, we achieved the sets $cpc(T)$, $cmb$, and $sepset$ including conditioning sets for all variables not in $cpc(T)$. Then in Algorithm 2, Steps 2 to 10 discover the spouses of $T$ as follows.

\begin{itemize}

\item $cpc(A)=\{E, T\}$, and $E\indep T|\{B, A\}$ in Figure~\ref{fig5-new2}(b) holds, $E\notin sp(T)$.

\item $cpc(B)=\{E, T\}$, and $B\in sepset(E)$ holds, $E\notin sp(T)$.

\item $cpc(G)=\{C, T\}$, and $C\notin cpc(T)$, $G\notin sepset(C)$ (i.e., $sepset(C)=\emptyset)$), then Algorithm 2 checks that $G\in cmb_b(T)$ and $C\nindep T|G$ in Figure~\ref{fig5-new2} (b). Thus, $C\in sp(T)$, and $cmb_b(T)=\{A, B, G, C\}$.
\end{itemize}

Thus, finally, by Algorithm 2, we get $mimb\_MB(T)=\bigcup_{i=1}^{3}cmb_i(T)=\{A, B, G, C\}$ and
$mimb\_pa(T)=\bigcap_{i=1}^{3}cmb_i(T)=\{A, B\}$.

\subsection{Discussion}

In the following, we discuss the correctness of the MIPC subroutine and  the MIMB algorithm, as summarised in Theorems~\ref{the5-23} and~\ref{the5-24} below.

\begin{theorem}
In Algorithm 3, when  $\Upsilon$ is conservative, the true set of parents and children of $T$ is a subset of the parent and children set found by Algorithm 3, i.e.  $pc(T)\subseteq cpc(T)$.

\emph{\textbf{Proof:}} First we prove that $cpc(T)$ includes $pc(T)$.  In the case of $\zeta_T=0$, by Lemma~\ref{lem4-11}, if $\forall V_j\in pc(T)$, then for $\forall D_i$,  $T\nindep V_j|S$ holds for $\forall S\subseteq cpc(T)\setminus V_j$. If $0<\zeta_T<n$, since  $\Upsilon$ is conservative, $\exists D_i$, $T\nindep V_j|S$ holds in $D_i$. Consequently, in Algorithm 3, $V_j$ enters $cpc(T)$ at Step 5 and and will remain in $cpc(T)$.

Second, assuming the set $nd(T)\setminus pa(T)$ denotes the non-descendants excluding the parents of $T$, we prove that $\forall V_j\in\{nd(T)\setminus pa(T)\}$ is not included in $cpc(T)$. By the Markov condition, without conditioning on $pa(T)$, some non-descendants of $T$ will enter $cpc(T)$ at Step 5. At Steps 11 and 27, since  $\Upsilon$ is conservative, $\exists D_i\in D$ such that $T$ is not manipulated, then $pa(T)\subseteq cmb_i(T)$, conditioning on $pa(T)$, $V_j$  is not in $cpc(T)$ in $D_i$.

Finally, we prove that some descendants of $T$ which are not in $ch(T)$ may be added to $cpc(T)$. By the Markov condition, given a descendant $X$ of $T$ and $pa(X)$, $T\indep X|pa(X)$. If $pa(X)\cap sp(T)\neq\emptyset$ holds, i.e., some variables in $pa(X)$ are spouses of $T$, we cannot find a dataset $D_i$ in $D$ satisfies $\{X\cup pa(X)\}\subseteq cmb_i(T)$ since Algorithm 3 cannot find any spouses of $T$.  By Corollary~\ref{cor5-2}, $X$ is not able to be removed at Steps 12 and 20.
\boxend
\label{the5-23}
\end{theorem}

Theorem~\ref{the5-23} concludes that all parents and children of $T$ will enter $cpc(T)$ in Algorithm 3, and sometimes may include some descendants of $T$ which are not $T$'s children.
For example, Figure~\ref{fig5-new1} illustrates the situation where some descendants of $T$ are added to $cpc(T)$. In Figure~\ref{fig5-new1}, $T$ is the target variable, $A$ is a spouse of $T$ in green which means that $A$ is manipulated, $B$ is a child of $T$, and $C$ is a descendant of $T$. $C$ will enter $cpc(T)$ and cannot be removed by the MIPC subroutine. This is because that $A\indep T$ holds such that $A$ is not added to $ipc(T)$ according to Algorithm 3, but only when conditioning on both $B$ and $A$, $T$, $C$ and $T$ are independent. Therefore after $C$ is added to $ipc(T)$, it is not removed and remains in $ipc(T)$.

\begin{figure}[t]
\centering
\includegraphics[height=1.2in,width=1.2in]{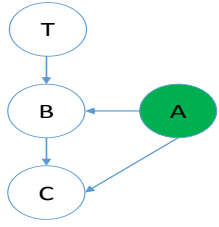}
\caption{An example where $cpc(T)$ in Algorithm 3 includes false positives}
\label{fig5-new1}
\end{figure}

To remove the false positives in the output of MIPC, we can employ a symmetry property in a DAG, that is, if $V_i\notin pc(V_j)$, $V_j\notin pc(V_i)$.  Theorem~\ref{the5-24} below describes that if symmetrical correction is applied to the output of the MIPC (Algorithm 3), the MIMB Algorithm (Algorithm 2) is theoretically sound.

\begin{theorem}
By a symmetry correction, in Algorithm 3, (1)  if $\zeta_T=0$ or $0<\zeta_T<n$, $\bigcup_{i=1}^{n}cmb_i(T)=MB(T)$; (2) $\bigcap_{i=1}^{n}cmb_i(T)=Pa(T)$ if $\zeta_T=0$ and $ch(T)\subset\bigcup_{i=1}^{n}\Upsilon_i$.

\textbf{Proof:} By the symmetry correction, (1) if $\zeta_T=0$ or $0<\zeta_T<n$ holds, by Theorems~\ref{the4-11} and~\ref{the4-21}, at Step  1 in Algorithm 2, $cmb_i(T)$ contains the true parents and children of $T$ in $D_i$. Then at Step 3 of Algorithm 2, the output is the true parents and children of each variable in $pc(T)$. By Theorem~\ref{the5-10}, at Steps 2 to 10 in Algorithm 2, the spouses of  $T$ in $D_i$ enter $cmb_i(T)$. Thus, $\bigcup_{i=1}^{n}cmb_i(T)=MB(T)$ holds. (2) By the proof in (1), for $\forall i\in\{1,\cdots,n\}$, $cmb_i(T)$ in $D_i$ is the true MB of $T$ in $D_i$, due to  $\zeta_T=0$ and $ch(T)\subset\bigcup_{i=1}^{n}\Upsilon_i$,  by Theorem~\ref{the4-14}, $\bigcap_{i=1}^{n}cmb_i(T)=pa(T)$ holds.
\boxend
\label{the5-24}
\end{theorem}

Since in practice the false positive as illustrated with the example in Figure~\ref{fig5-new1}, similar to most algorithms for MB discovery from a single observational dataset~\citep{aliferis2010local1}, in our implementation, symmetry correction is not applied to the output of MIPC.

\subsection{Complexity of MIMB and the baseline algorithm}


Using the number of independence tests for measuring time complexity, in the MIPC algorithm (Algorithm 3), at Steps 2 to 10, the complexity of checking variables in $V\setminus T$ is $O(n|V|)$.  From Steps 11 to 27, MIPC examines the subsets only containing the newly added features at Steps 12 to 19.  Assuming the largest examined subset size within $ipc(T)$ is up to $\ell$ at Steps 11 to 27 in Algorithm 3, the complexity of MIPC is $O(|cpc(T)||\ell^22^\ell)$ where $2^\ell$  considers those subsets in $ipc(T)$ that only contain the newly added variables. Thus, for a single dataset, the complexity of MIMB is 
$O(|cpc(T)|\ell^22^\ell)$.  Assuming $k$ is the average number of datasets examined by Algorithm 3 for discovering $cpc(T)$, in the best case of $k=1$, the complexity of MIMB is$O(|cpc(T)|\ell^22^\ell)$, while in the average case of $k<n$, the complexity of MIMB is $O(k|cpc(T)|\ell^22^\ell)$. In the worst case of $k=n$, the complexity of MIMB is $O(n|cpc(T)|\ell^22^\ell)$.

For the baseline algorithm, since it employs  the existing HITON-MB algorithm for MB discovery from each dataset. For a single dataset, the complexity of HITON-MB is $O(|V|\ell^22^{|ipc(T)|})$ where $2^{|ipc(T)|}$  includes all subsets in the set $ipc(T)$ with the largest size~\citep{aliferis2010local1}. Then the time complexity of the baseline algorithm is $O(n|V|\ell^22^{|ipc(T)|})$. Thus, in the average case, MIMB is more efficient than the baseline algorithm, while  in the worst case where MIMB needs to check all datasets, the time complexity of MIMB may approximate to that of the baseline algorithm.

\section{Experiments}\label{sec6}

In this section, we evaluate the proposed MIMB algorithm. For the evaluation, we compare the performance of MIMB with the baseline algorithm described in section 5.1, as well as the He-Geng algorithm~\citep{he2016causal}.
As there are no algorithms specifically developed for finding MBs from multiple interventional datasets, the He-Geng algorithm, which learns an entire DAG from multiple interventional datasets, becomes the only option for our comparative studies.  We run the He-Geng algorithm to learn an entire DAG from a dataset to obtain the MB of a target variable from the learnt DAG, then compare the MB with the MB of the target found by MIMB and the baseline algorithm.

In the experiments, we apply a series of synthetic data sets and a real-world data set for evaluating the baseline algorithm, MIMB, and the He-Geng algorithm. $G^2$ tests  are used for all the conditional independence tests and the significance level, $\alpha$ for the $G^2$ test is set to 0.01.

\subsection{Experiments on Synthetic Data}

With the synthetic data, we evaluate and compare the performance of the three algorithms  using the following metrics:
\begin{itemize}
\item  Precision. The number of true positives in the output (i.e. the variables in the output belonging to the true MB of a target variable) divided by the number of variables in the output (the MB found) by an algorithm. 

\item Recall. The number of true positives in the output divided by the number of variables included in the true MB of a target variable. 

\item F1 score. $\small F1=2*(precision*recall)/(precision+recall)$.

\item nTest. The number of conditional independence tests for the MB discovery implemented by an algorithm.
\end{itemize}

We  conduct two simulations to generate two types of multiple interventional datasets using a commonly used benchmark Bayesian network, the 37-variable ALARM (A Logical Alarm Reduction Mechanism) network\footnote{Refer to www.bnlearn.com/bnrepository for the details of the network.}, as shown in Figure~\ref{newfig6_1}.
The first simulation implements five intervention experiments, which generate five interventional datasets, while the second simulation implements ten intervention experiments for the generation of ten interventional datasets. Each dataset contains 5000 samples. 

For the experiments, we run each of the two simulations for 10 times to generate 10 groups of the 5 datasets with the first simulation (and denote this collection of 50 datasets as ``nData=5"), and 10 groups of the 10 datasets with the second simulation (and denote this collection of 100 datasets as ``nData=10")
We compute the average  precision, recall, F1 score, and nTest for each algorithm over the ten groups of datasets produced by the two types of simulations, respectively.

\begin{figure}[t]
\centering
\includegraphics[height=3in,width=3.2in]{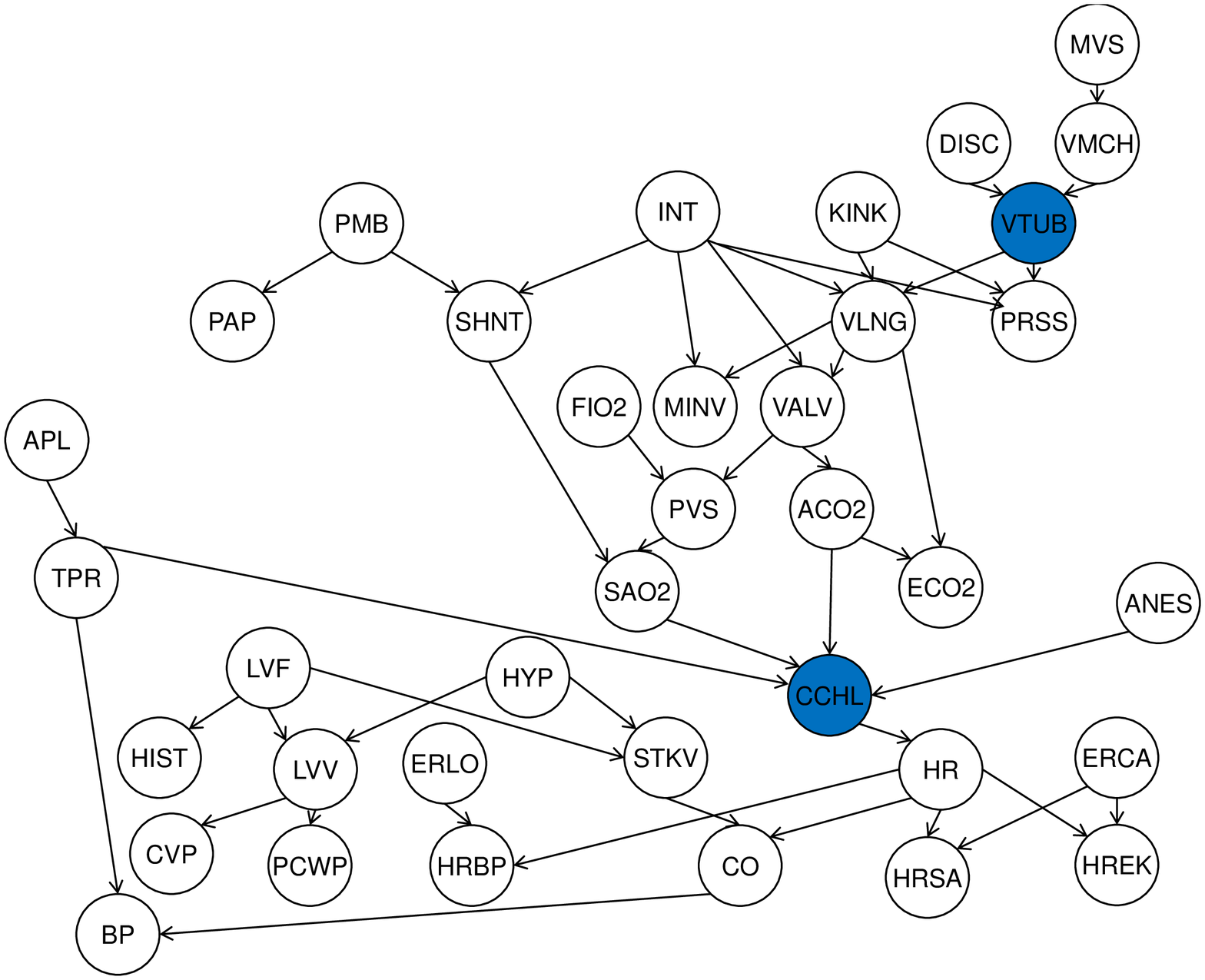}
\caption{The ALARM Bayesian network}
\label{newfig6_1}
\end{figure}

Referring to Figure~\ref{newfig6_1}, in both simulations, we choose the variables  ``VTUB" and  ``CCHL"  (i.e. the two blue nodes) in the ALARM network as the target variables, respectively.  ``VTUB"  has the largest sized MB among all variables in the network while  ``CCHL"  has the largest parent set and the second largest MB comparing to other variables in the network.

Given a target variable, in an intervention experiment in a simulation, 
the manipulated variables are randomly chosen, and we make sure the multiple (5 for the first simulation and 10 for the second simulation) intervention experiments are conservative.
After the manipulated variables are chosen in an experiment, by the derived the post-intervention DAG, the post-intervention conditional probabilities of each manipulated variable are then generated from an uninformative Dirichlet distribution~\citep{he2016causal}. Based on post-intervention conditional probabilities and the structure of the post-intervention DAG, we generate interventional datasets. By the analyses in Section~\ref{sec4}, in the simulation experiments, $T$ is manipulated less than $n$ times. In Tables~\ref{tb6-1} to~\ref{tb6-21} as follows, $A\pm B$ denotes that the average performance measure (precision, recall or F1) is A with a standard deviation of B.
The best results are highlighted in bold-face.
\begin{table}[t]
\centering
\caption{Results of discovering the MB of  ``VTUB" ($\zeta_T=0$)}
\label{tb6-1}
\begin{tabular}{|l|c|c|l|}
\hline
\multirow{2}{*}{Algorithm} & \multicolumn{3}{c|}{nData=5 ($\zeta_T=0$)}                                                            \\ \cline{2-4}
                           & recall                               & precision                            & \multicolumn{1}{c|}{F1} \\ \hline
He-Geng                    & 0.6333$\pm$0.07                      & 0.9800$\pm$0.06                      & 0.7661$\pm$0.06         \\ \hline
Baseline                   & \textbf{1.00$\pm$0.00}              & \textbf{1.00$\pm$0.00}              & \textbf{1.00$\pm$0.00}          \\ \hline
MIMB                       & \textbf{1.00$\pm$0.00}              & \textbf{1.00$\pm$0.00}              & \textbf{1.00$\pm$0.00}           \\ \hline
\multicolumn{1}{|c|}{}     & \multicolumn{3}{c|}{nData=10 ($\zeta_T=0$)}                                                           \\ \hline
He-Geng                    & \multicolumn{1}{l|}{0.667$\pm$0.00}  & \multicolumn{1}{l|}{\textbf{1.00$\pm$0.00}}   & 0.80$\pm$0.00           \\ \hline
Baseline                   & \multicolumn{1}{l|}{\textbf{1.00$\pm$0.00}}   & \multicolumn{1}{l|}{0.9847$\pm$0.05} & \textbf{0.9923$\pm$0.02}  \\ \hline
MIMB                       & \multicolumn{1}{l|}{0.9667$\pm$0.07} & \multicolumn{1}{l|}{\textbf{1.00$\pm$0.00}}   & 0.9818$\pm$0.04         \\ \hline
\end{tabular}
\end{table}
\begin{table}[t]
\centering
\caption{Results of  discovering the MB of  ``VTUB" ($0<\zeta_T<n$)}
\label{tb6-11}
\begin{tabular}{|l|c|c|l|}
\hline
\multirow{2}{*}{Algorithm} & \multicolumn{3}{c|}{nData=5 }                                                         \\ \cline{2-4}
                           & recall                              & precision                            & \multicolumn{1}{c|}{F1} \\ \hline
He-Geng                    & 0.60$\pm$0.09                       & \textbf{1.00$\pm$0.00}                        & 0.7467$\pm$0.07         \\ \hline
Baseline                      & \textbf{0.9667$\pm$0.07}                     & 0.9286$\pm$0.08                      & 0.9434$\pm$0.04         \\ \hline
MIMB                       & \textbf{0.9667$\pm$0.07}                    & 0.9429$\pm$0.07                      & \textbf{0.9510$\pm$0.04}         \\ \hline
\multicolumn{1}{|c|}{}     & \multicolumn{3}{c|}{nData=10}                                                          \\ \hline
He-Geng                    & \multicolumn{1}{l|}{0.667$\pm$0.08} & \multicolumn{1}{l|}{\textbf{1.00$\pm$0.00}}   & 0.7967$\pm$0.06         \\ \hline
Baseline                     & \multicolumn{1}{l|}{\textbf{1.00$\pm$0.00}}  & \multicolumn{1}{l|}{0.9095$\pm$0.11} & \textbf{0.9492$\pm$0.07}  \\ \hline
MIMB                       & \multicolumn{1}{l|}{0.95$\pm$0.08}  & \multicolumn{1}{l|}{\textbf{0.9381$\pm$0.08}} & 0.9421$\pm$0.06         \\ \hline
\end{tabular}
\end{table}
\begin{table}[t]
\centering
\caption{Results of discovering the MB  of  ``CCHL" ($\zeta_T=0$)}
\label{tb6-2}
\begin{tabular}{|l|c|c|l|}
\hline
\multirow{2}{*}{Algorithm} & \multicolumn{3}{c|}{nData=5}                                                                        \\ \cline{2-4}
                           & recall                             & precision                            & \multicolumn{1}{c|}{F1} \\ \hline
He-Geng                    & 0.70$\pm$0.11                      & \textbf{1.00$\pm$0.00}                        & 0.8194$\pm$0.07         \\ \hline
Baseline                   & \textbf{1.00$\pm$0.00}                     & 0.8262$\pm$0.08                      & 0.9030$\pm$0.05         \\ \hline
MIMB                       & 0.92$\pm$0.10                      & 0.9514$\pm$0.10                      & \textbf{0.9300$\pm$0.08}         \\ \hline
\multicolumn{1}{|c|}{}     & \multicolumn{3}{c|}{nData=10}                                                                       \\ \hline
He-Geng                    & \multicolumn{1}{l|}{0.80$\pm$0.00} & \multicolumn{1}{l|}{0.98$\pm$0.06}   & 0.8800$\pm$0.03         \\ \hline
Baseline                 & \multicolumn{1}{l|}{\textbf{1.00$\pm$0.00}} & \multicolumn{1}{l|}{0.7976$\pm$0.06} & 0.8864$\pm$0.04         \\ \hline
MIMB                       & \multicolumn{1}{l|}{0.86$\pm$0.14} & \multicolumn{1}{l|}{\textbf{1.00$\pm$0.00}}   & \textbf{0.9194$\pm$0.08}         \\ \hline
\end{tabular}
\end{table}
\begin{table}[t]
\centering
\caption{Results of discovering the MB of  ``CCHL" ($0<\zeta_T<n$)}
\label{tb6-21}
\begin{tabular}{|l|c|c|l|}
\hline
\multirow{2}{*}{Algorithm} & \multicolumn{3}{c|}{nData=5}                                                                        \\ \cline{2-4}
                           & recall                             & precision                            & \multicolumn{1}{c|}{F1} \\ \hline
He-Geng                    & 0.66$\pm$0.10                      & \textbf{1.00$\pm$0.00}                        & 0.7917$\pm$0.07         \\ \hline
Baseline                  & \textbf{1.00$\pm$0.00}                      & 0.8667$\pm$0.07                      & 0.9273$\pm$0.04         \\ \hline
MIMB                       & 0.94$\pm$0.10                      & \textbf{1.00$\pm$0.00}                        & \textbf{0.9667$\pm$0.05}         \\ \hline
\multicolumn{1}{|c|}{}     & \multicolumn{3}{c|}{nData=10}                                                                       \\ \hline
He-Geng                    & \multicolumn{1}{l|}{0.70$\pm$0.11} & \multicolumn{1}{l|}{0.95$\pm$0.16}   & 0.7990$\pm$0.11         \\ \hline
Baseline                  & \multicolumn{1}{l|}{\textbf{1.00$\pm$0.00}} & \multicolumn{1}{l|}{0.8298$\pm$0.14} & \textbf{0.9008$\pm$0.09} \\ \hline
MIMB                       & \multicolumn{1}{l|}{0.84$\pm$0.08} & \multicolumn{1}{l|}{\textbf{0.96$\pm$0.12}}   & 0.8661$\pm$0.08         \\ \hline
\end{tabular}
\end{table}

\subsubsection{Recall, Precision, and F1}

\textbf{MIMB and the baseline vs. the He-Geng algorithm.} From Tables~\ref{tb6-1} to~\ref{tb6-21}, both  the baseline and MIMB algorithms are significantly better on the recall  and F1 metrics than the He-Geng algorithm all the time.   The baseline and MIMB find much more true positives than the He-Geng algorithm.  The recall metric determines whether an algorithm is able to find a correct MB of a target variable. For example, except for the results in Table~\ref{tb6-11}, the recall of the baseline is up to 1. The He-Geng algorithm is better than the baseline and MIMB on the precision metric under certain conditions. The explanation for the better precision is that the He-Gang algorithm implicitly applies symmetry corrections. The He-Geng algorithm needs to find the neighbors of all variables for learning the entire structure. If variable $X$ is not adjacent to variable $Y$, the He-Geng algorithm will do not include $X$ in the neighbor set of $Y$. However, in the experiments, both MIMB and the baseline do not implement symmetry correction.

\textbf{MIMB vs. the baseline.} On the recall metric, from Tables~\ref{tb6-1} to~\ref{tb6-21},  the baseline achieves the highest recall values (up to 1 at most times). MIMB is little inferior to the baseline on the recall metric. This is because  the baseline uses the union of the MBs found in each dataset separately as the final MB of a target variable. 
For MIMB, if a variable is not in the MB found in one dataset, the algorithm does not test this variable any more for its membership in the MB (Steps 11 to 27 in the MIPC algorithm).
Thus, the problem is that when a variable is mistakenly disregarded due to data noise or sample bias of the dataset, MIMB will not add the variable to the final MB, thus a false negative. But when a false positive is added to the MB found in a dataset by  the baseline, then the false positive cannot be removed from the output of  the baseline. So this also explains why MIMB is better than  the baseline on the precision metric. Thus,  the baseline has a better performance than MIMB on the recall metric while MIMB is superior to  the baseline on the precision metric, thus, the F1 values of the two algorithms are very competitive.

\subsubsection{Efficiency of the three algorithms}

We use the number of independence  tests carried out by an algorithm as the measure of its efficiency. Tables~\ref{tb6-new11} and~\ref{tb6-new12} show that under all conditions, 
MIMB conducts much fewer tests than both the baseline and the He-Geng algorithm. 
The He-Geng algorithm is slower than MIMB because it needs to learn an entire DAG containing all variables involved in a dataset in order to get the MB of a target variable. Although not learning an entire DAG, the baseline needs to perform the same independence tests in each dataset. MIMB avoids the unnecessary tests in all datasets.

\begin{table}[t]
\centering
\caption{Number of independence tests of finding the MB of  ``VTUB"}
\label{tb6-new11}
\begin{tabular}{|l|l|l|l|l|}
\hline
\multirow{2}{*}{Algorithm} & \multicolumn{2}{c|}{$\zeta_T=0$} & \multicolumn{2}{c|}{$0<\zeta_T<n$} \\ \cline{2-5} 
                           & nData=5       & nData=10       & nData=5       & nData=10       \\ \hline
He-Geng   &28,375$\pm$2280               &55,287$\pm$4347                  & 24,574$\pm$3231             & 53,525$\pm$2705 \\ \hline

Baseline    &2,584$\pm$126                   &3,483$\pm$518                     &1,308$\pm$432     &3,174$\pm$675                   \\ \hline

MIMB      &\textbf{1,102$\pm$85}          & \textbf{1,843$\pm$210}       & \textbf{922$\pm$204}        & \textbf{1,738$\pm$184}        \\ \hline
\end{tabular}
\end{table}

\begin{table}[t]
\centering
\caption{Number of independence tests of finding the MB of  ``CCHL"}
\label{tb6-new12}
\begin{tabular}{|l|r|r|r|r|}
\hline
\multirow{2}{*}{Algorithm} & \multicolumn{2}{c|}{$\zeta_T=0$} & \multicolumn{2}{c|}{$0<\zeta_T<n$} \\ \cline{2-5} 
                           & nData=5       & nData=10       & nData=5       & nData=10       \\ \hline
He-Geng                    &28,342$\pm$2248                &54,454$\pm$3959                 &24,714$\pm$3336               & 52,042$\pm$3010                \\ \hline

Baseline                    &2,568$\pm$303                  & 4,983$\pm$528                  & 1,805$\pm$486                       &3,837$\pm$598         \\ \hline
MIMB                      &\textbf{1,390$\pm$196} & \textbf{2,400$\pm$747}      & \textbf{1,332$\pm$179}          & \textbf{2,166$\pm$307}                   \\ \hline
\end{tabular}
\end{table}




\begin{table}[t]
\centering
\caption{Results  of  discovering the MB  of ``VTUB" with different $\alpha$ ($\zeta_T=0, nData=10$)}
\label{tb6-5}
\begin{tabular}{|l|c|c|c|c|}
\hline
\multirow{2}{*}{Algorithm} & \multicolumn{4}{c|}{$\alpha$=0.01}                                                                                                         \\ \cline{2-5}
                           & recall                             & precision                          & F1                                   & nTest                                \\ \hline
He-Geng                    & 0.67$\pm$0.00                      & 1.00$\pm$0.00                      & 0.80$\pm$0.00                        & 55,287$\pm$4347                      \\ \hline
Baseline                      & 1.00$\pm$0.00                      & 0.98$\pm$0.05                      & 0.9923$\pm$0.02                      & 3,483$\pm$518                        \\ \hline
MIMB                       & 0.97$\pm$0.07                      & 1.00$\pm$0.00                      & 0.9818$\pm$0.06                      & 1,843$\pm$210                        \\ \hline
                           & \multicolumn{4}{c|}{$\alpha$=0.05}                                                                                                         \\ \hline
He-Geng                    & \multicolumn{1}{l|}{0.67$\pm$0.00} & \multicolumn{1}{l|}{1.00$\pm$0.00} & \multicolumn{1}{l|}{0.80$\pm$0.00}   & \multicolumn{1}{l|}{63,289$\pm$3853} \\ \hline
Baseline                      & \multicolumn{1}{l|}{1.00$\pm$0.00} & \multicolumn{1}{l|}{0.85$\pm$0.12} & \multicolumn{1}{l|}{0.9141$\pm$0.07} & \multicolumn{1}{l|}{3,853$\pm$477}   \\ \hline
MIMB                       & \multicolumn{1}{l|}{0.97$\pm$0.07} & \multicolumn{1}{l|}{0.98$\pm$0.05} & \multicolumn{1}{l|}{0.9742$\pm$0.06} & \multicolumn{1}{l|}{2,163$\pm$284}   \\ \hline
\end{tabular}
\end{table}
\begin{table}[t]
\centering
\caption{Results of  discovering the MB  of ``VTUB" with different $\alpha$ ($0<\zeta_T<n, nData=10$)}
\label{tb6-6}
\begin{tabular}{|l|l|l|l|l|}
\hline
\multirow{2}{*}{Algorithm} & \multicolumn{4}{c|}{ $\alpha$=0.01}                                                                                     \\ \cline{2-5}
                           & \multicolumn{1}{c|}{recall} & \multicolumn{1}{c|}{precision} & \multicolumn{1}{c|}{F1} & \multicolumn{1}{c|}{nTest} \\ \hline
He-Geng                    & 0.67$\pm$0.08               & 1.00$\pm$0.00                  & 0.7976$\pm$0.06         & 53,525$\pm$2705            \\ \hline
Baseline                      & 1.00$\pm$0.00               & 0.91$\pm$0.11                  & 0.9492$\pm$0.07         & 3,174$\pm$675              \\ \hline
MIMB                       & 0.95$\pm$0.08               & 0.94$\pm$0.08                  & 0.9421$\pm$0.06         & 1,738$\pm$184              \\ \hline
                           & \multicolumn{4}{c|}{ $\alpha$=0.05}                                                                                     \\ \hline
He-Geng                    & 0.68$\pm$0.09               & 0.95$\pm$0.11                  & 0.7876$\pm$0.07         & 62,228$\pm$2608            \\ \hline
Baseline                      & 1.00$\pm$0.00               & 0.71$\pm$0.14                  & 0.8199$\pm$0.09         & 3,821$\pm$777              \\ \hline
MIMB                       & 0.95$\pm$0.08               & 0.93$\pm$0.10                  & 0.9358$\pm$0.08         & 1,968$\pm$298              \\ \hline
\end{tabular}
\end{table}

\begin{table}[t]
\centering
\caption{Results of  discovering the MB  of  ``CCHL'' with different $\alpha$ ($T=0$, nData=10)}
\label{tb6-61}
\begin{tabular}{|l|c|c|c|c|}
\hline
\multirow{2}{*}{Algorithm} & \multicolumn{4}{c|}{ $\alpha$=0.01}                                    \\ \cline{2-5} 
                           & recall           & precision     & F1            & nTest           \\ \hline
He-Geng                    & 0.80$\pm$0.00    & 0.98$\pm$0.06 & 0.88$\pm$0.03 & 54,454$\pm$3959 \\ \hline
Baseline                   & 1.00$\pm$0.00    & 0.80$\pm$0.06 & 0.89$\pm$0.04 & 4,983$\pm$525   \\ \hline
MIMB                       & 0.86$\pm$0.14    & 1.00$\pm$0.00 & 0.92$\pm$0.08 & 2,400$\pm$747   \\ \hline
                           & \multicolumn{4}{c|}{ $\alpha$=0.05}                                    \\ \hline
He-Geng                    & 0.80$\pm$0.00    & 0.98$\pm$0.06 & 0.88$\pm$0.03 & 62,608$\pm$4489 \\ \hline
Baseline                   & 1.00$\pm$0.00 & 0.71$\pm$0.11 & 0.81$\pm$0.07 & 5,530$\pm$618   \\ \hline
MIMB                       & 0.90$\pm$0.11    & 0.96$\pm$0.07 & 0.93$\pm$0.07 & 2,999$\pm$644   \\ \hline
\end{tabular}
\end{table}

\begin{table}[t]
\centering
\caption{Results of  discovering the MB  of  ``CCHL'' with different $\alpha$ ($0<\zeta_T<n$, nData=10)}
\label{tb6-62}
\begin{tabular}{|l|c|c|c|c|}
\hline
\multirow{2}{*}{Algorithm} & \multicolumn{4}{c|}{ $\alpha$=0.01}                                 \\ \cline{2-5} 
                           & recall        & precision     & F1            & nTest           \\ \hline
He-Geng                    & 0.70$\pm$0.11 & 0.95$\pm$0.16 & 0.80$\pm$0.11 & 52,042$\pm$3010 \\ \hline
Baseline                   & 1.00$\pm$0.00 & 0.83$\pm$0.14 & 0.90$\pm$0.09 & 3,837$\pm$598   \\ \hline
MIMB                       & 0.84$\pm$0.10 & 0.96$\pm$0.12 & 0.89$\pm$0.08 & 2,166$\pm$307   \\ \hline
                           & \multicolumn{4}{c|}{ $\alpha$=0.05}                                 \\ \hline
He-Geng                    & 0.80$\pm$0.13 & 0.92$\pm$0.13 & 0.84$\pm$0.06 & 61,841$\pm$9378 \\ \hline
Baseline                   & 1.00$\pm$0.00 & 0.71$\pm$0.08 & 0.83$\pm$0.05 & 4,237$\pm$567   \\ \hline
MIMB                       & 0.90$\pm$0.11 & 0.94$\pm$0.10 & 0.92$\pm$0.09 & 2,827$\pm$442   \\ \hline
\end{tabular}
\end{table}

\subsubsection{Impact of parameter $\alpha$}

We use the results of the ``nData=10" datasets to illustrate the impact of parameter $\alpha$ on the three algorithms for MB discovery. Considering ``VTUB" as the target variable, Tables~\ref{tb6-5} and~\ref{tb6-6} show that $\alpha$, the significance level for conditional independence tests, has little influence on the He-Geng and MIMB algorithms using the recall, precision, and F1 metrics.

With ``CCHL" as the target, Tables~\ref{tb6-61} and~\ref{tb6-62} show that as the value of $\alpha$ changes from 0.01 to 0.05, under the condition of $T=0$ (i.e. number of interventions on $T$ up to 0), the He-Geng algorithm has no changes on the recall, precision, and F1 metrics, while with $0<\zeta_T<n$, the recall values of the He-Geng algorithm changes from 0.70 to 0.80.  For the baseline, for $T=0$ or $0<\zeta_T<n$,  it gets the same recall values under different values of $\alpha$. But the precision values of the baseline have a significant change using different values of $\alpha$. In contrast, the value of $\alpha$ has less impact on MIMB than the baseline and the He-Geng algorithm.

The explanation is that the baseline simply uses the union of the MBs found in different datasets separately. By the union, the baseline will make more true positives enter the final output, but the baseline does not attempt to remove the false positives from its output. Therefore, the precision of the baseline decreases as the value of $\alpha$ increases.  As we discussed previously, the He-Geng algorithm implements a symmetry correction to remove false positives, while MIMB leverages the information of multiple datasets as much as possible to identify false positives. Thus, both MIMB and the He-Geng algorithm achieve stable precision than the baseline.
Additionally, the three algorithms all conducted more tests when $\alpha=0.01$ than when $\alpha=0.05$.

\subsubsection{The discovery of parents}

When $\zeta_T=0$ and $ch(T)\subseteq\bigcup_{i=1}^{n}\Upsilon_i$ hold, by Theorem~\ref{the4-14} in Section 4.1, $\bigcap_{i=1}^{n}MB_i(T)$ equals to $pa(T)$.  Tables~\ref{tb6-new4} to~\ref{tb6-new51} report the results of $\bigcap_{i=1}^{n}MB_i(T)$ produced by the baseline and MIMB using the ``nData=5" and ``nData=10" datasets, respectively, for different $\alpha$ values. Meanwhile, in Tables~\ref{tb6-new4} to~\ref{tb6-new51}, for the He-Geng algorithm, since it combines the learnt DAGs from each dataset to form a final DAG, we uses the parents of a given target by the union of parents of the target in each found DAG learnt from multiple datasets. 

From Table 14, when $\alpha=0.01$,  with ``nData=5",  both the baseline and MIMB find all parents of  ``VTUB". With ``nData=10", the He-Geng algorithm finds all parents of  ``VTUB" without any false positives, while the baseline and MIMB do not find all parents.  The output of MIMB does not include any false positives. 
 
When $\alpha=0.05$, Table~\ref{tb6-new41} shows that with  ``nData=5",  MIMB achieves much better recall and F1 values than the baseline and the He-Geng algorithm.
With ``nData=10", the He-Geng algorithm finds the exact set of parents of  ``VTUB", , and MIMB still performs better than the baseline.
 
Table~\ref{tb6-new5} shows the results on  ``CCHL".  When $\alpha=0.01$,  with  ``nData=5", both MIMB, all three algorithms have  achieved 100\% precision. On the recall value, MIMB is better than the baseline and the He-Geng algorithm. With ``nData=10",  the He-Geng algorithm finds the exact set of parents of  ``CCHL", while MIMB and the baseline obtain almost the same recall, precision, and F1 values.

When $\alpha=0.05$, Table~\ref{tb6-new51} shows that MIMB and the He-Geng algorithm are very competitive, and the baseline has the worst result with the ``nData=5".  With ``nData=10", the He-Geng algorithm finds the exact set of parents of   ``CCHL'', and MIMB's performance is much better than the baseline.

Why does the He-Geng algorithm have the best performance with ``nData=10" for finding the parent sets?  The explanation is that the He-Geng algorithm first finds an entire DAG in each dataset, then it obtains the parents of a given target by taking the union of parents of the target in each found DAG. On the other hand, MIMB and the baseline discover the parents of a given target by taking the intersection of found MBs of the target in different datasets (MIMB does not go through EACH dataset). If the faithfulness assumption holds and all tests are reliable, when $\zeta=0$, $\bigcap_{i=1}^{n}MB_i(T)$ returned by MIMB and the baseline should equal to $pa(T)$.  But in practice, due to noise in data and the violation of the faithfulness assumption, the intersection may not be equal to $pa(T)$.  For example, assuming $X$ is a parent of $T$, in $D_j$, the baseline and MIMB are able to add $X$ to $MB_j(T)$, but in $D_i$, they may not. Thus, finally $\bigcap_{i=1}^{n}MB_i(T)$ will not include $X$.

\begin{table}[t]
\centering
\caption{Results of $\bigcap_{i=1}^{n}MB_i(T)$ for ``VTUB" ($\alpha$=0.01)}
\label{tb6-new4}
\begin{tabular}{|l|c|c|l|}
\hline
\multirow{2}{*}{Algorithm} & \multicolumn{3}{c|}{nData=5}                                                                      \\ \cline{2-4}
                           & recall                             & precision                          & \multicolumn{1}{c|}{F1} \\ \hline
 He-Geng                     & 0.80$\pm$0.42                      & 1.00$\pm$0.00                      & 0.89$\pm$0.32           \\ \hline
                           
Baseline                      & 1.00$\pm$0.00                      & 1.00$\pm$0.00                      & \textbf{1.00$\pm$0.00}           \\ \hline
MIMB                       & 1.00$\pm$0.00                      & 1.00$\pm$0.00                      & \textbf{1.00$\pm$0.00}           \\ \hline
                           & \multicolumn{3}{c|}{nData=10}                                                                     \\ \hline
He-Geng                     & \multicolumn{1}{l|}{1.00$\pm$0.00 } & \multicolumn{1}{l|}{1.00$\pm$0.00 } & \textbf{1.00$\pm$0.00}          \\ \hline                     
Baseline                         & \multicolumn{1}{l|}{0.90$\pm$0.21} & \multicolumn{1}{l|}{0.93$\pm$0.14} & 0.8933$\pm$0.15         \\ \hline
MIMB                       & \multicolumn{1}{l|}{0.90$\pm$0.21} & \multicolumn{1}{l|}{1.00$\pm$0.00} & 0.9333$\pm$0.14         \\ \hline
\end{tabular}
\end{table}

\begin{table}[t]
\centering
\caption{Results of $\bigcap_{i=1}^{n}MB_i(T)$ for ``VTUB" ($\alpha$=0.05)}
\label{tb6-new41}
\begin{tabular}{|l|c|c|l|}
\hline
\multirow{2}{*}{Algorithm} & \multicolumn{3}{c|}{nData=5}                                                                      \\ \cline{2-4}
                           & recall                             & precision                          & \multicolumn{1}{c|}{F1} \\ \hline
 He-Geng                     & 0.70$\pm$0.48                      & 1.00$\pm$0.00                      & 0.70$\pm$0.46           \\ \hline
                           
Baseline                      & 0.85$\pm$0.24                      & 0.65$\pm$0.19                      & 0.7233$\pm$0.19           \\ \hline
MIMB                       & 0.90$\pm$0.21                      & 0.8667$\pm$0.17                      & \textbf{0.8533$\pm$0.14}           \\ \hline
                           & \multicolumn{3}{c|}{nData=10}                                                                     \\ \hline
He-Geng                     & \multicolumn{1}{l|}{1.00$\pm$0.00 } & \multicolumn{1}{l|}{1.00$\pm$0.00 } & \textbf{1.00$\pm$0.00}          \\ \hline                     
Baseline                         & \multicolumn{1}{l|}{0.85$\pm$0.24} & \multicolumn{1}{l|}{0.9833$\pm$0.05} & 0.88$\pm$0.16         \\ \hline
MIMB                       & \multicolumn{1}{l|}{0.90$\pm$0.21} & \multicolumn{1}{l|}{1.00$\pm$0.00} & 0.9333$\pm$0.14         \\ \hline
\end{tabular}
\end{table}

\begin{table}[t]
\centering
\caption{Results of $\bigcap_{i=1}^{n}MB_i(T)$ for ``CCHL" ($\alpha$=0.01)}
\label{tb6-new5}
\begin{tabular}{|l|l|l|l|}
\hline
\multicolumn{1}{|c|}{\multirow{2}{*}{Algorithm}} & \multicolumn{3}{c|}{nData=5}                                                           \\ \cline{2-4}
\multicolumn{1}{|c|}{}                           & \multicolumn{1}{c|}{recall} & \multicolumn{1}{c|}{precision} & \multicolumn{1}{c|}{F1} \\ \hline
He-Geng                                          & 0.875$\pm$0.13               & 1.00$\pm$0.00                      & 0.9286$\pm$0.08         \\ \hline

Baseline                                             & 0.90$\pm$0.13               & 1.00$\pm$0.00                      & \textbf{0.9429$\pm$0.07}         \\ \hline
MIMB                                             & 0.90$\pm$0.17               & 1.00$\pm$0.00                      & 0.9381$\pm$0.11      \\ \hline
\multicolumn{1}{|c|}{}                           & \multicolumn{3}{c|}{nData=10}                                                          \\ \hline

He-Geng                                           & 1.00$\pm$0.00              & 1.00$\pm$0.00                     & \textbf{1.00$\pm$0.00}        \\ \hline
Baseline                                             & 0.925$\pm$0.17              & 1.00$\pm$0.00                     & 0.9571$\pm$0.07         \\ \hline
MIMB                                             & 0.926$\pm$0.22              & 1.00$\pm$0.00                     & 0.9571$\pm$0.07         \\ \hline
\end{tabular}
\end{table}

\begin{table}[t]
\centering
\caption{Results of $\bigcap_{i=1}^{n}MB_i(T)$ for  ``CCHL" ($\alpha$=0.05)}
\label{tb6-new51}
\begin{tabular}{|l|l|l|l|}
\hline
\multicolumn{1}{|c|}{\multirow{2}{*}{Algorithm}} & \multicolumn{3}{c|}{nData=5}                                                           \\ \cline{2-4}
\multicolumn{1}{|c|}{}                           & \multicolumn{1}{c|}{recall} & \multicolumn{1}{c|}{precision} & \multicolumn{1}{c|}{F1} \\ \hline
He-Geng                                          & 0.90$\pm$0.12               & 0.98$\pm$0.06                      & \textbf{0.9317$\pm$0.07}        \\ \hline

Baseline                                             & 0.80$\pm$0.10               & 0.98$\pm$0.08
 & 0.8750$\pm$0.07         \\ \hline
MIMB                                             & 0.875$\pm$0.13               & 1.0$\pm$0.00                      & 0.9286$\pm$0.08         \\ \hline
\multicolumn{1}{|c|}{}                           & \multicolumn{3}{c|}{nData=10}                                                          \\ \hline

He-Geng                                           & 1.00$\pm$0.00              & 1.00$\pm$0.00                     & \textbf{1.00$\pm$0.00}         \\ \hline
Baseline                                             & 0.80$\pm$0.16              & 1.00$\pm$0.00                     & 0.8810$\pm$0.10         \\ \hline
MIMB                                             & 0.926$\pm$0.22              & 1.00$\pm$0.00                     & 0.9571$\pm$0.07         \\ \hline
\end{tabular}
\end{table}

\subsection{Experiments on Real-world Data}

In the section, we use the real-world data set about educational attainment of teenagers provided in~\citep{rouse1995democratization,stock2003introduction} as a possible practical application of the MIMB algorithm. The original data set includes records of 4739 pupils from approximately 1100 US high schools and 14 attributes as shown in Table~\ref{tb6-new1}.

Following the method in~\citep{peters2016causal},  variable {\em distance} is  the manipulated variable, and split the original data set into two interventional data sets (for which the {\em distance} variable is intervened): one includes 2231 data instances of all pupils who live closer to a 4-year college than the median distance of 10 miles, and the other includes 2508 data instances of all pupils who live at least 10 miles from the nearest 4-year college. Then we select the variable {\em education} as the target variable and  make it  into a binary target, that is, whether a pupil received a BA (Bachelor of Arts) degree or not.
\begin{table}[t]
\centering
\caption{Variables in the educational attainment  data set and their meanings}
\begin{tabular}{|l|l|}
\hline
\textbf{Variable} & \textbf{Meaning} \\ \hline
education & \begin{tabular}[c]{@{}l@{}}Years of education completed (target variable, binarized to completed\\ a BA or not in this paper)\end{tabular}\\ \hline
gender & Student gender, male or female \\ \hline
ethnicity & Afam/Hispanic/Other \\ \hline
score & \begin{tabular}[c]{@{}l@{}}Base year composite test score. (These are achievement tests given to\\high school seniors in the sample)\end{tabular} \\ \hline
fcollege & Father is a college graduate or not\\ \hline
mcollege & Mother is a colllege graduate or not \\ \hline
home & Family owns a house or not \\ \hline
urban & School in urban area or not \\ \hline
unemp & County unempolyment rate in 1980 \\ \hline
wage & State hourly wage in manufacturing in 1980 \\ \hline
distance & Distance to the nearest 4-year college \\ \hline
tuition & Avg. state 4-year college tuition in \$1000's \\ \hline
income & Family income \textgreater \$25,000 per year or not \\ \hline
region & Student in the western states or other states \\ \hline
\end{tabular}
\label{tb6-new1}
\end{table}

As we have not the ground truth of the causes and effects of the variables in this real-world data set, we use the PC algorithm~\citep{spirtes2000causation}, a well-known algorithm for Bayesian network structure learning to learn a partial DAG (see Figure~\ref{fig6_new1}) from the original dataset. This causal structure is then used as the ground truth in our experiments. 
The work in~\citep{peters2016causal} also applied their proposed method, ICP (Invariant Causal prediction, details see Section 2) to the two interventional datasets (created from the educational attainment dataset as described above) to find the causes of the variable {\em education}. Therefore, in our experiments with the real-world data for MB and cause (parent) discovery, we also compare MIMB with ICP, in addition to the baseline and the He-Geng algorithm.

Table~\ref{tb6-new0} shows that MIMB is more efficient than He-Geng and the baseline, with much fewer conditional independence tests done. Meanwhile, considering Figure~\ref{fig6_new1} as the ground truth, all the four parents (causes) of {\em education} discovered by MIMB  is consistent with the parents of {\em education} in Figure~\ref{fig6_new1}. But ICP  finds only two parents, while He-Geng and the baseline only discover three correct parents each. 

To further validate those results, Table~\ref{tb6-new2} gives the p-values of the strength of influences of the variables on {\em education} calculated by each algorithm.  By Table~\ref{tb6-new2}, all the four algorithms show that {\em score} and {\em fcollege} have the most significant influences on {\em education}. Meanwhile, MIMB shows that  {\em income}
and {\em mcollege} are more important than {\em tuition}, which seems plausible.
\begin{figure}[t]
\centering
\includegraphics[height=2.2in,width=3in]{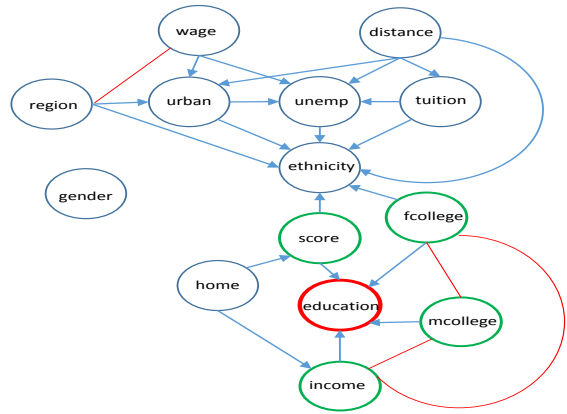}
\caption{A causal structure learned from the original educational attainment data set (red lines indicate the edges whose directions are undetermined by PC)}
\label{fig6_new1}
\end{figure}
 \begin{table*}[t]
\centering
\caption{The causes and MBs of {\em education} discovered by each algorithm and nTest (Number of tests)}
\begin{tabular}{|c|c|c|c|c|}
\hline
       & ICP             & He-Geng                                                                              & Baseline                                                                                             & MIMB                                                                                         \\ \hline
Causes & score, fcollege & \begin{tabular}[c]{@{}c@{}}score, fcollege, \\income, tuition                                                    \end{tabular} & \begin{tabular}[c]{@{}c@{}}score, fcollege,\\ income \end{tabular}  &\begin{tabular}[c]{@{}c@{}} score, fcollege,\\ mcollege, income \end{tabular}                                                           \\ \hline
MBs    & -               & \begin{tabular}[c]{@{}c@{}}score, fcollege,\\ mcollege, income,\\ tuition\end{tabular} & \begin{tabular}[c]{@{}c@{}}score, fcollege,\\ mcollege, income, \\ tuition, ethnicity,\end{tabular} & \begin{tabular}[c]{@{}c@{}}score, fcollege,\\ mcollege, income,\\ tuition, region\end{tabular} \\ \hline
nTest  & -               & 29,895                                                                                & 1,075                                                                                              & 491                                                                                          \\ \hline
\end{tabular}
\label{tb6-new0}
\end{table*}
\begin{table}[t]
\centering
\caption{p-values of influences each variable on {\em education} (``$\bullet$" denotes that the corresponding variable has a significant influences on {\em education})}
\begin{tabular}{|l|l|l|l|l|}
\hline
          & ICP            & He-Geng          & Baseline            & MIMB             \\ \hline
gender    & 0.187          & 0.6941           & 0.6941           & 0.6941           \\ \hline
ethnicity & 0.167          & 2.2E-04$\bullet$ & 0.0041           & 0.0067           \\ \hline
score     & 0.031$\bullet$ & 1.3E-06$\bullet$ & 3.3E-07$\bullet$ & 2.7E-07$\bullet$ \\ \hline
fcollege  & 0.096$\bullet$ & 1.5E-05$\bullet$ & 6.5E-07$\bullet$ & 6.1E-07$\bullet$ \\ \hline
mcollege  & 0.189          & 2.9E-04$\bullet$ & 8.9E-04$\bullet$ & 3.3E-04$\bullet$ \\ \hline
home      & 0.213          & 0.0114           & 0.0178           & 0.0058           \\ \hline
urban     & 0.163          & 0.0487           & 0.2186           & 0.1347           \\ \hline
unemp     & 0.213          & 0.7365           & 0.6711           & 0.8182           \\ \hline
wage      & 0.180          & 0.5265           & 0.4206           & 0.3787           \\ \hline
tuition   & 0.213          & 2.0E-04$\bullet$ & 7.5E-04$\bullet$ & 0.0068           \\ \hline
income    & 0.151          & 4.0E-05$\bullet$ & 4.2E-04$\bullet$ & 5.9E-04$\bullet$ \\ \hline
region    & 0.208          & 0.0116           & 0.1588           & 0.0065           \\ \hline
\end{tabular}
\label{tb6-new2}
\end{table}

In Table~\ref{tb6-new2},  we can see that the MBs found by the three algorithms (He-Geng, baseline, and MIMB) are a little different. We should be aware that without knowing the ``real" ground-truth of the MB of {\em education}, it is difficult to tell which algorithm discovers the correct MB of {\em education}, although we have a reference causal structure in Figure~\ref{fig6_new1}. 

However, as the MB of a target variable is the set of optimal feature for classification on the target~\citep{aliferis2010local2}, we  evaluate the findings of MIMB by examining the performance of the predictions based on the MBs found from the multiple interventional data sets. With KNN and NB (Naive Bayes) classifiers, we use those discovered MBs for predicting the target {\em education}, that is, whether a student will receive a BA degree or not.

Firstly, we  select 2000 data instances from the two interventional data sets respectively to construct two training data sets and the remaining 739 data instances as the testing data set. The training datasets and the testing dataset created in this way will have non-identical distribution, thus posing challenges on predictions. 
Secondly, we use He-Geng, the baseline and MIMB to discover the MBs of  {\em education} from the two training data sets. Thirdly, in each of the two training data sets, we train the KNN and NB classifiers using the discovered MBs and make predictions on the testing dataset. For each algorithm, we combine the prediction results of the KNN and NB  classifiers on testing data by majority voting. We repeat the experiments ten times and report the average classification accuracy and the number of tests, as shown in Table~\ref{tb6_new3}.

From the table, we see that MIMB achieves higher classification accuracy than both He-Geng and the baseline, and MIMB is significantly more efficient than He-Geng and the baseline.
\begin{table}[t]
\centering
\caption{Classification accuracy and number of tests (A$\pm$ B denotes that A is the average
classification accuracy and B is the standard deviation)}
\begin{tabular}{|l|l|l|l|}
\hline
 &He-Geng &Baseline &MIMB \\ \hline
NB & 0.7428$\pm$0.0127 & 0.7461$\pm$0.0136 & \textbf{0.7494$\pm$0.0173} \\ \hline
KNN &  0.7201$\pm$0.0272 & 0.6855$\pm$0.0392 & \textbf{0.7225$\pm$0.0368} \\ \hline
nTest &24877$\pm$2609 & 2498$\pm$996 & \textbf{317$\pm$105} \\ \hline
\end{tabular}
\label{tb6_new3}
\end{table}

\section{Conclusion and future work}\label{sec7}

In the paper, we have studied the problem of discovering MBs from multiple interventional datasets without knowing which variables were manipulated. From this study, we can see that multiple interventional data are useful and can be beneficial to MB discovery. The work in this paper is the first to present the theorems about the conditions for the discovery and and the algorithm (MIMB) to find the MBs and and the parent set of a given target variable under the conditions. Using sythetic and real-world datasets, experimental results validate the theorems and the MIMB algorithm proposed in the paper. In future, we will explore if the theorems and MIMB can be utilized to improve global causal structure discovery with multiple interventional datasets.

\vskip 0.2in

\bibliography{sigproc}

\end{sloppy}
\end{document}